\documentclass[letterpaper,twocolumn,10pt]{article}
\usepackage[available,functional, reproduced]{usenixbadges}
\usepackage{usenix}

\usepackage{tikz}
\usepackage{amsmath}
\usepackage[ruled,vlined,linesnumbered]{algorithm2e}

\SetCommentSty{mycommfont}
\usepackage{xcolor}
\usepackage{color}
\definecolor{codegreen}{rgb}{0,0.6,0}
\definecolor{codegray}{rgb}{0.5,0.5,0.5}
\definecolor{codepurple}{rgb}{0.58,0,0.82}
\definecolor{backcolour}{rgb}{0.95,0.95,0.92}
\definecolor{textblue}{rgb}{.2,.2,.7}
\definecolor{textred}{rgb}{0.54,0,0}
\definecolor{textgreen}{rgb}{0,0.43,0}
\definecolor{codered}{rgb}{201,72,12}
\usepackage[T1]{fontenc}
\usepackage[scaled=0.85]{beramono} 
\usepackage{listings}
\usepackage{multirow}
\usepackage{multicol}
\usepackage{url}
\PassOptionsToPackage{hyphens}{url}
\usepackage{hyperref}
\newcommand\code[1]{\texttt{\textcolor{black}{#1}}}

\lstdefinestyle{tt}{
language=Python,
basicstyle=\linespread{0.9}\ttfamily\footnotesize,
breaklines=true,
numbers=left,
frame=single,
numberstyle=\tiny, 
stepnumber=1,
numbersep=5pt, 
tabsize=4,
keywordstyle=\bfseries\color{codegreen},
commentstyle=\color{textred},   
stringstyle=\color{textgreen},
columns=fullflexible,
keepspaces=true,
xleftmargin=\parindent,
showstringspaces=false,
otherkeywords = {True, False},
keywordstyle=[2]\color{codepurple}\bfseries,
keywords=[2]{TCG, TC_GNN},
keywordstyle=[3]\color{textblue}\bfseries,
keywords=[3]{__init__, forward},
keywordstyle=[4]\color{codegreen},
keywords=[4]{self},
}
\lstdefinestyle{tt1}{
language=Python,
basicstyle=\linespread{0.9}\ttfamily\footnotesize,
breaklines=true,
numbers=left,
frame=single,
numberstyle=\tiny, 
stepnumber=1,
numbersep=5pt, 
tabsize=4,
keywordstyle=\bfseries\color{codegreen},
commentstyle=\color{textred},   
stringstyle=\color{textgreen},
columns=fullflexible,
keepspaces=true,
xleftmargin=\parindent,
showstringspaces=false,
otherkeywords = {True, False},
keywordstyle=[2]\color{codepurple}\bfseries,
keywords=[2]{wmma},
keywordstyle=[3]\color{textblue}\bfseries,
keywords=[3]{fragment, load_matrix_sync, mma_sync, store_matrix_sync},
keywordstyle=[4]\color{codegreen}\bfseries,
keywords=[4]{matrix_a, row_major, column_major, mem_row_major, tf32},
}
\lstdefinestyle{tt2}{
language=Python,
basicstyle=\linespread{0.9}\ttfamily\footnotesize,
breaklines=true,
numbers=left,
frame=single,
numberstyle=\tiny, 
stepnumber=1,
numbersep=5pt, 
tabsize=4,
keywordstyle=\bfseries\color{codegreen},
commentstyle=\color{textred},   
stringstyle=\color{textgreen},
columns=fullflexible,
keepspaces=true,
xleftmargin=\parindent,
showstringspaces=false,
otherkeywords = {True, False},
keywordstyle=[2]\color{codepurple}\bfseries,
keywords=[2]{wmma},
keywordstyle=[3]\color{textblue}\bfseries,
keywords=[3]{fragment, load_matrix_sync, mma_sync, store_matrix_sync},
keywordstyle=[4]\color{codegreen}\bfseries,
keywords=[4]{matrix_a, row_major, column_major, mem_row_major, tf32},
}
\usepackage{soul,listings,xcolor}
\lstdefinestyle{tt3}{
language=C,
basicstyle=\linespread{0.9}\ttfamily\footnotesize,
breaklines=true,
numbers=left,
frame=single,
numberstyle=\tiny, 
stepnumber=1,
numbersep=5pt, 
tabsize=4,
keywordstyle=\bfseries\color{codegreen},
commentstyle=\color{textred},   
stringstyle=\color{textgreen},
columns=fullflexible,
keepspaces=true,
xleftmargin=\parindent,
showstringspaces=false,
otherkeywords = {True, False},
keywordstyle=[2]\color{codepurple}\bfseries,
keywords=[2]{BLK_H, BLK_W, wmma},
keywordstyle=[3]\color{textblue}\bfseries,
keywords=[3]{fragment, load_matrix_sync, mma_sync, store_matrix_sync, __shared__},
keywordstyle=[4]\color{codegreen}\bfseries,
keywords=[4]{matrix_a, row_major, column_major, mem_row_major, tf32},
}

\lstdefinestyle{tt4}{
language=Python,
basicstyle=\linespread{0.9}\ttfamily\footnotesize,
breaklines=true,
numbers=left,
frame=single,
numberstyle=\tiny, 
stepnumber=1,
numbersep=5pt, 
tabsize=4,
keywordstyle=\bfseries\color{codegreen},
commentstyle=\color{textred},   
stringstyle=\color{textgreen},
columns=fullflexible,
keepspaces=true,
xleftmargin=\parindent,
showstringspaces=false,
otherkeywords = {True, False},
keywordstyle=[2]\color{codepurple}\bfseries,
keywords=[2]{TCGNN, TCGNN},
keywordstyle=[3]\color{textblue}\bfseries,
keywords=[3]{__init__, forward},
keywordstyle=[4]\color{codegreen},
keywords=[4]{self},
}
\usepackage{array}
\usepackage{setspace}
\usepackage{subfig}
\usepackage{ctable} 
\newcommand\subfig[1]{\textcolor{green}{#1}}

\SetAlFnt{\small}
\SetAlCapFnt{\small}
\newcommand\hlp[1]{\underline{\textbf{\textit{{#1}}}}}

\newcommand*{\Mname}{TC-GNN}
\begin{document}

\date{}
\pagestyle{empty}

\title{\Large \bf TC-GNN: Bridging Sparse GNN Computation and Dense Tensor Cores on GPUs}

\author{
{\rm Yuke Wang, Boyuan Feng, Zheng Wang, Guyue Huang, and Yufei Ding}\\
{\rm \textit{University of California, Santa Barbara}}
}

\maketitle

\begin{abstract}
Recently, graph neural networks (GNNs), as the backbone of graph-based machine learning, demonstrate great success in various domains (\textit{e.g.}, e-commerce). However, the performance of GNNs is usually unsatisfactory due to the highly sparse and irregular graph-based operations. To this end, we propose \textbf{TC-GNN}
\footnote{Paper is accepted to USENIX ATC'23.}
, the first GNN acceleration framework based on GPU Tensor Core Units (TCUs).
The core idea is to reconcile the ``Sparse'' GNN computation with the high-performance ``Dense'' TCUs. 
Specifically, we conduct an in-depth analysis of the sparse operations in mainstream GNN computing frameworks.
We introduce a novel sparse graph translation technique to facilitate TCU processing of the sparse GNN workload.
We implement an effective CUDA core and TCU collaboration design to fully utilize GPU resources.
We integrate \Mname~with the PyTorch framework for high programmability.
Rigorous experiments show an average of $1.70\times$ speedup over the state-of-the-art DGL framework across various models and datasets.
%
\end{abstract}
\section{Introduction} \label{sect: introduction}
Over the recent years, with the increasing popularity of graph-based learning, graph neural networks (GNNs)~\cite{GCNConv, GINConv, AGNN} become dominant in the computing of essential tasks across a wide range of domains, like e-commerce, financial services, and etc.
Compared with standard methods for graph analytics, such as random walk~\cite{grover2016node2vec, deepWalk, huang2021broader} and graph laplacians~\cite{luo2011cauchy, luo2009non, cheng2018deep}, GNNs highlight themselves with significantly higher accuracy~\cite{GCNConv, GINConv, GATConv} and better generality~\cite{SageConv}.
From the computation perspective, GNNs feature an interleaved execution phase of both graph operations (scatter-and-gather~\cite{gonzalez2012powergraph}) at the \textit{\textbf{Aggregation}} phase and Neural Network (NN) operations (matrix multiplication) at the \textit{\textbf{Update}} phase. Our experimental studies further show that the aggregation phase which involves highly sparse computation on irregular input graphs generally takes more than 80\% of the running time for both GNN training and inference. 
Existing GNN frameworks, \textit{e.g.}, Deep Graph Library~\cite{wang2019dgl} and PyTorch Geometric~\cite{pyG}, are mostly built upon the popular NN frameworks that are originally optimized for dense operations, such as general matrix-matrix multiplication (GEMM). 
To support sparse computations in GNNs, their common strategy is to incorporate sparse primitives (such as cuSPARSE~\cite{cusparse}) for their backend implementations. 
However, cuSPARSE leverages the sparse linear algebra (LA) algorithm which involves lots of high-cost indirect memory accesses on non-zero elements of a sparse matrix. Therefore, cuSPARSE cannot enjoy the same level of optimizations (e.g., data reuse) as its dense counterpart, such as cuBLAS~\cite{cublas}. Moreover, cuSPARSE is designed to only utilize CUDA cores. Therefore, It cannot benefit from advancements in GPU hardware features, like Tensor Core Units (TCUs) on the recent NVIDIA Ampere and Hopper GPUs. 
Such a design is also the trend of many other AI-tailored accelerators/units (e.g., Google TPU~\cite{jouppi2017datacenter} and Matrix Core~\cite{matrixcore} on AMD GPUs) and can significantly boost the performance of dense LA algorithms (e.g., GEMM and Convolution) in most conventional deep-learning applications (e.g., CV~\cite{he2016deep} and NLP~\cite{devlin2018bert}).

This work focuses on exploring the potential of TCUs for accelerating such GNN-based graph learning and our design/optimization principles will also benefit other similar AI hardware~\cite{jouppi2017datacenter, matrixcore} for sparse deep-learning workloads.
We remark that making TCUs effective for general GNN computing is a non-trivial task. 
Our initial study shows that naively applying the TCU to sparse GNN computation would even result in inferior performance compared with the existing sparse implementations on CUDA cores.
There are several challenges. 
\hlp{First}, directly resolving the sparse GNN computing problem with the pure dense GEMM solution is impractical due to the extremely large memory cost ($\mathcal{O}(N^2)$, where $N$ is the number of nodes). 
Besides, traversing the matrix tiles already known to be filled with all-zero elements would cause excessive unnecessary computations and memory access.
\hlp{Second}, simply employing TCUs to process non-zero matrix tiles of the sparse graph adjacency matrix would still waste most of the TCU computation and memory access efforts. 
This is because TCU input matrix tiles are defined with fixed dimension settings (\textit{e.g.}, $\mathit{height} (16)\times \mathit{width}(8)$), whereas the non-zero elements of a sparse graph adjacency matrix are distributed irregularly. Thus, it requires intensive zero-value padding to satisfy such a rigid input constraint.
\hlp{Third}, although the recent CUDA release update enables TCUs to exploit the benefit of certain types of sparsity~\cite{blocked_spmm}, it only supports 
blocked SpMM, where non-zero elements must first fit into well-shaped blocks and the number of blocks must be the same across different rows.
Such an input restriction makes it hard to handle highly irregular sparse graphs in real-world GNN applications.

To this end, we introduce, \textbf{\Mname}\footnote{\url{https://github.com/YukeWang96/TC-GNN_ATC23.git}}, the first TCU-based GNN acceleration design on GPUs. 
Our key insight is to \textit{let the sparse input graph fit the dense computation of TCUs}. 
\textit{\textbf{At the input level}}, instead of exhaustively traversing all sparse matrix tiles and determining whether to process each tile, we develop a new \textit{sparse graph translation} (SGT) technique that can effectively identify those non-zero tiles and condense non-zero elements from these tiles into fewer number of ``dense'' tiles. 
Our major observation is that neighbor sharing is very common among nodes in real-world graphs. Therefore, applying SGT can effectively merge the unnecessary data loading of the shared neighbors among different nodes to avoid high-cost memory access.
SGT is generic to any kind of sparse pattern of input graphs and can always yield the correct results as the original sparse algorithm.
\textit{\textbf{At the GPU kernel level}}, for efficiently processing GNN sparse workloads, \Mname~exploits the benefits of CUDA core and TCU collaboration. The major design idea is that the CUDA core, which is more powerful at fine-grained thread-level execution, would be a good candidate for managing memory-intensive data access. While TCU, which is more powerful in handling simple arithmetic operations (\textit{e.g.}, multiplication and addition), would be well-suited for compute-intensive GEMM on dense tiles generated from SGT.
\textit{\textbf{At the framework level}}, we integrate \Mname~with the popular PyTorch~\cite{pytorch} framework. Thereby, users only need to interact with their familiar PyTorch programming environment by using \Mname~APIs. This can significantly reduce extra learning efforts, and improve user productivity and code portability. 

To sum up, we summarize our contributions as follows:
\begin{itemize}
    \item We conduct a detailed analysis ($\S$\ref{sect: Motivation}) of existing solutions (\textit{e.g.}, SpMM on CUDA cores) and identify the potentials of TCUs for accelerating sparse GNN workloads.
    \item We introduce a sparse graph translation technique ($\S$\ref{sect: TCU-Aware Sparse Graph Translation}). It can make the sparse and irregular GNN input graphs easily fit the dense computing of TCUs for acceleration.
    \item We build a TCU-tailored algorithm ($\S$\ref{sect: TCU-tailored GNN Computation}) and GPU kernel design ($\S$\ref{sect: TCU-centric Workload Mapping}) for CUDA core and TCU collaboration on GPUs to handle different sparse GNN computation. 
    %
    %
    \item Extensive experiments show \Mname~achieves 1.70$\times$ speedup on average over the state-of-the-art GNN computing framework, Deep Graph Library, across various mainstream GNN models and dataset settings. 
\end{itemize}

\section{Background}
\label{sect: background and related work}
\subsection{Graph Neural Networks}
\label{sect: Graph Neural Networks}
\begin{figure} [t] \small
    \centering
    \includegraphics[width=0.95\columnwidth]{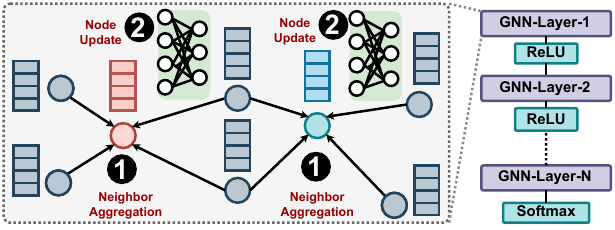}
    \vspace{-5pt}
    \caption{GNN General Computation Flow.}
    \label{fig: GNN Computation Flow}
     \vspace{-10pt}
\end{figure}
Graph neural networks (GNNs) are an effective tool for graph-based machine learning. The detailed computing flow of GNNs is illustrated in Figure~\ref{fig: GNN Computation Flow}. 
GNNs basically compute the node feature vector (embedding) for node $v$ at layer $k+1$ based on the embedding information at layer $k$ ($k \geq 0$), as shown in Equation~\ref{eq: GNN},
\begin{gather} \small \label{eq: GNN}
 \begin{aligned} 
   a_{v}^{(k+1)}  &= \boldsymbol{\mathit{Aggregate}}^{(k+1)}({h_{u}^{(k)}|u\in \mathbf{N}(v)\cup h_v^{(k)}}) \\
   h_{v}^{(k+1)}  &= \boldsymbol{\mathit{Update}}^{(k+1)}(a_{v}^{(k+1)})
\end{aligned}   
\end{gather}
where $h_{v}^{(k)}$ is the embedding vector for node $v$ at layer $k$; $a_{v}^{(k+1)}$ is the aggregation results through collecting neighbors' information (\textit{e.g.}, node embeddings); $\mathbf{N}(v)$ is the neighbor set of node $v$.
The aggregation method and the order of aggregation and update could vary across different GNNs. 
Some methods~\cite{GCNConv, SageConv} just rely on the neighboring nodes while others~\cite{GATConv} also leverage the edge properties that are computed by applying vector dot-product between source and destination node embeddings. The update function is generally composed of standard NN operations, such as a fully connected layer or a multi-layer perceptron (MLP) in the form of $w\cdot a_{v}^{(k+1)} + b$, where $w$ and $b$ are the weight and bias parameters, respectively. The common choices for node embedding dimensions are 16, 64, and 128, and the embedding dimension may change across different layers. 
After several iterations of aggregation and update (\textit{i.e.}, several GNN layers), we will get the output feature embedding of each node, which can be used for various downstream graph learning tasks, such as node classification~\cite{kaspar2010graph, gibert2012graph, duran2017learning} and link prediction~\cite{chen2005link, kunegis2009learning, tylenda2009towards}. 
 \begin{figure} [t] \small
    \centering
    \includegraphics[width=\columnwidth]{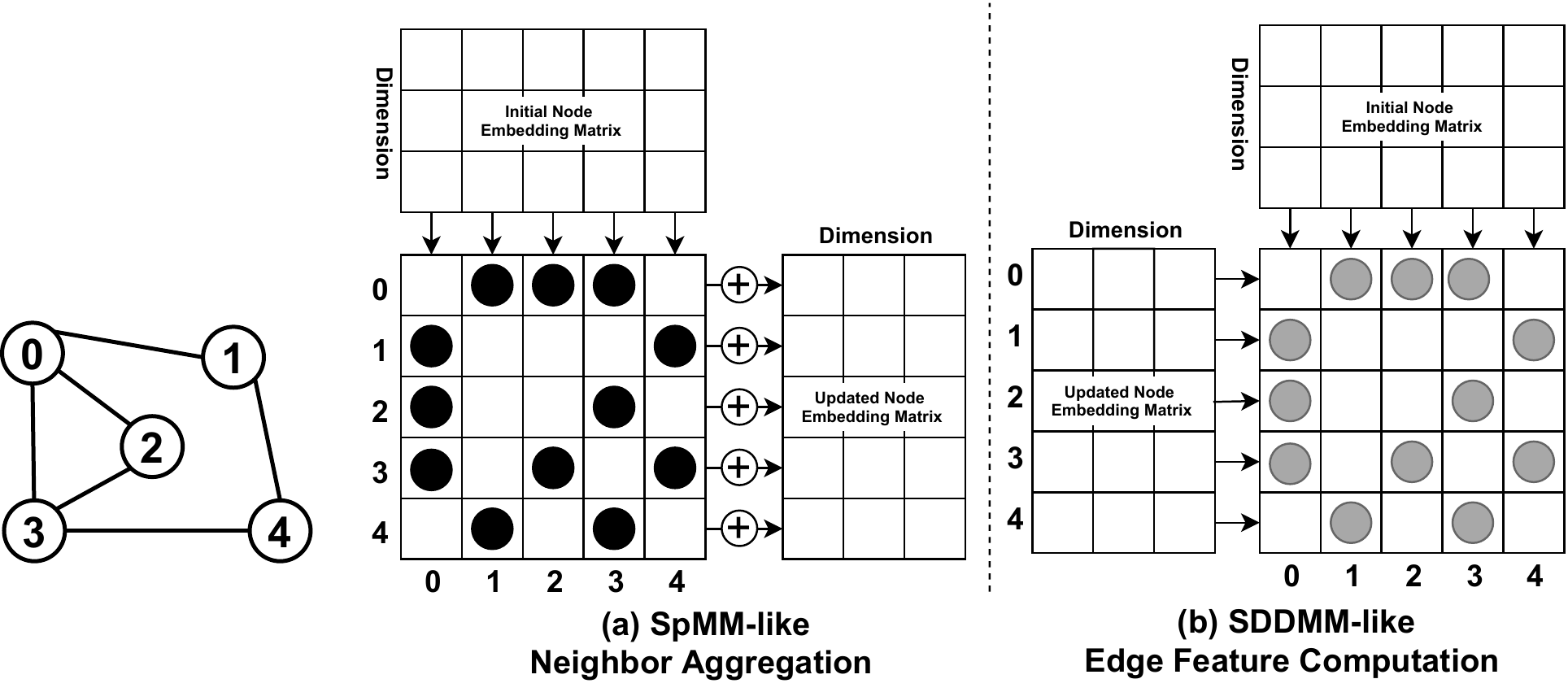}
    \vspace{-10pt}
    \caption{(a) SpMM-like and (b) SDDMM-like Operation in GNNs. 
    Note that ``$\rightarrow$'' indicates loading data; ``$\oplus$'' indicates neighbor embedding accumulation.}
    \label{fig: SpMM and SDDMM}
     \vspace{-10pt}
\end{figure}

The sparse computing in the aggregation phase is generally formalized as the sparse-matrix dense-matrix multiplication (SpMM), as illustrated in Figure~\ref{fig: SpMM and SDDMM}\subfig{a}, and is handled by many sparse libraries (\textit{e.g.}, cuSPARSE~\cite{cusparse}) in many state-of-the-art GNN frameworks~\cite{wang2019dgl, GNNAdvisor}. 
These designs only count on GPU CUDA cores for computing, which waste the modern GPUs with diverse computing units, such as the Tensor Core Unit (TCU). 
Specifically, we formalized the neighbor aggregation as SpMM-like operations (Equation~\ref{equ: SpMM})
\begin{equation}\small \label{equ: SpMM}
    \mathbf{\hat{X}} = (\mathbf{F}_{N\times N} \odot \mathbf{A}_{N\times N}) \cdot X_{N\times D})
\end{equation}
where $\mathbf{A}$ is the graph adjacency matrix stored in CSR format. 
$\mathbf{X}$ is a node feature embedding matrix stored in dense format. 
$N$ is the number of nodes in the graph, and $D$ is the size of node feature embedding dimension; 
$\odot$ is the elementwise multiplication and $\cdot$ is the standard matrix-matrix multiplication; 
$\mathbf{F}$ is the edge feature matrix in CSR format and can be computed by Sampled Dense-Dense Matrix Multiplication (SDDMM)-like operations (Equation~\ref{equ: SDDMM} and Figure~\ref{fig: SpMM and SDDMM}\subfig{b}). 
\begin{equation} \small \label{equ: SDDMM}
    \mathbf{F} = (\mathbf{X}_{N\times D} \cdot \mathbf{X}^{T}_{N\times D}) \odot \mathbf{A}_{N\times N} 
\end{equation}
Note that the computation of $F$ is optional in GNNs, which is generally adopted by the Attention-based Graph Neural Network in PyTorch~\cite{AGNN} for identifying more complicated graph structural information. 
Other GNNs, such as the Graph Convolutional Network~\cite{GCNConv} and Graph Isomorphism Network~\cite{GINConv}, only use the adjacency matrix for neighbor aggregation. 

\subsection{GPU Tensor Core}
\begin{figure} [t] \small
    \centering
\includegraphics[width=1\columnwidth]{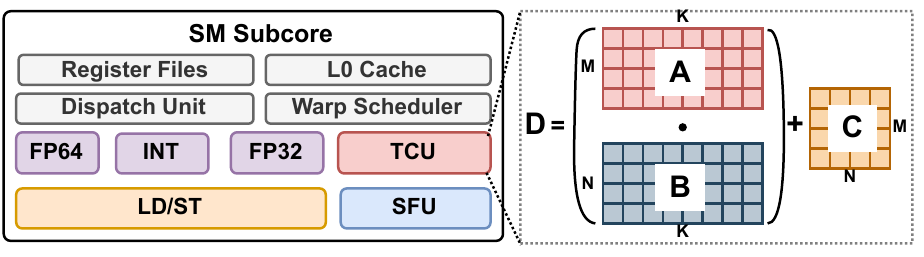}
    \vspace{-15pt}
    \caption{A Subcore of GPU SM with TCUs.}
    \label{fig: GPU TCU}
     \vspace{-15pt}
\end{figure}

In the most recent GPU architectures (since Volta~\cite{volta}), NVIDIA announced a new type of computing unit, Tensor Core Unit (TCU), for accelerating dense deep-learning operations (\textit{e.g.}, Dense GEMM). A GPU Streaming-Multiprocessor (w/ TCU) is illustrated in Figure~\ref{fig: GPU TCU}. Note that FP64, FP32, INT, and SFU are for double-precision, single-precision, integer, and special function units, respectively. 
Different from scalar computation on CUDA cores, TCU provides tile-based matrix-matrix computation primitives on register fragments, which can deliver more than $10\times$ throughput improvement.
In particular, TCU supports the compute primitive of $\mathbf{D} = \mathbf{A}\cdot \mathbf{B} + \mathbf{C}$, where $\mathbf{A}$ and $\mathbf{B}$ are required to be a certain type of precision (\textit{e.g.}, half, TF-32), while $\mathbf{C}$ and $\mathbf{D}$ are stored in FP32. 
Depending on the data precision and GPU architecture version, the matrix size (MMA shape) of $\mathbf{A}(M\times K)$, $\mathbf{B}(K\times N)$, and $\mathbf{C}(M\times N)$ should follow some principles~\cite{TC-rules}. 
For example, TF-32 TCU computing requires $M=N=16$ and $K=8$.
In the recent CUDA release (>=$11.0$) on Ampere ($sm$>=$80$), TF-32 serves as a good alternative to float/double on TCU-based GPU computing for modern deep-learning applications, according to NVIDIA's in-depth studies~\cite{TCcore-tf32}.

Different from the CUDA cores that operate at the thread level (\textit{e.g.}, allowing the ``if'' branch among threads), TCU supports only the operation at the warp level (\textit{e.g.}, forbidding the ``if'' branch among threads within a warp).
Before calling TCUs, all registers in a warp need to collaboratively store matrix tiles into a new memory hierarchy \textit{Fragment}~\cite{TCProgram}, which allows data sharing across registers. This intra-warp sharing provides opportunities for fragment-based memory optimizations.
TCU can be utilized in several ways. The simplest way is to call cuBLAS~\cite{cublas} by using the \texttt{cublasSgemmEX} API. 
The second way is to call the Warp Matrix Multiply-Accumulate (WMMA) (\texttt{nvcuda::wmma}) API~\cite{wmma} in CUDA C to operate TCUs directly with four major operations (Listing~\ref{code: wmma API interface.}). 
\begin{figure}[t] \small
\begin{lstlisting}[caption=WMMA APIs for TCUs in CUDA C.,style=tt3,label={code: wmma API interface.}]
wmma::fragment<matrix_a, M, N, K, tf32, row_major> a_frag;
// Load tiles (global/shared mem. -> register fragments).
wmma::load_matrix_sync(a_frag, A, M);
// Execute GEMM on loaded tiles on register fragments.
wmma::mma_sync(c_frag, a_frag, b_frag, c_frag);
// Move results (register fragments -> global/shared mem).
wmma::store_matrix_sync(C, c_frag, N, mem_row_major);
\end{lstlisting} 
\vspace{-10pt}
\end{figure}

Since the appearance of the TCU, research efforts have been devoted to accelerating deep-learning (DL) workloads with TCUs. 
%
Ang and Simon~\cite{Li2020AcceleratingBN} leverage 1-bit GEMM capability on Turing TCUs for accelerating binary Neural Network inference.
Boyuan et al.~\cite{EMM-TC} introduce GEMM-based scientific computing on TCUs with extended precision and high performance.
Yuke et al.~\cite{wang2022qgtc} treat batched quantized GNNs (partitioning large graphs into small graphs as batches) as batched dense GEMM computation and accelerate it on TCUs for inference.
These prior efforts use TCUs in the dense DL applications that TCU is initially designed for, while {\Mname} jumps out of the scope defined by TCU designers and accelerates the sparse full-graph GNNs using TCUs. 
\section{Motivation}
\label{sect: Motivation}
In this section, we will discuss the major technical thrust for us to leverage TCUs for accelerating sparse GNN computation. We use the optimization of SpMM as the major example in this discussion, and the acceleration of SDDMM would also benefit from similar optimization principles.

\subsection{SpMM on CUDA cores} \label{sect: SpMM on CUDA cores}
As the major component of sparse linear algebra operation, SpMM has been incorporated in many off-the-shelf libraries~\cite{cusparse,lapack99,magma,intel-alt, ge-spmm}. 
The close-sourced cuSPARSE~\cite{cusparse} library developed by NVIDIA is the most popular solution and it can deliver state-of-the-art performance for most GPU-based SpMM computation.
cuSPARSE has also been widely adopted by many GNN frameworks, such as Deep Graph Library (DGL)~\cite{wang2019dgl}, as the backend for sparse operations.
To understand its characters, we profile DGL on one layer of a GCN~\cite{GCNConv} model (\textit{neighbor aggregation} + \textit{node update}) on NVIDIA RTX3090. We report two key kernel matrices for only neighbor aggregation kernel, including L1/texture cache hit rate ($Cache$) and the achieved Streaming-Multiprocessor (SM) occupancy ($Occ.$). We select three representative GNN datasets: Cora with 3,327 nodes, 9,464 edges, and 3,703 node embedding dimensions; Citeseer with 2,708 nodes, 10,858 edges, and 1,433 dimensions; Pubmed with 19,717 nodes, 88,676 edges, and 500 dimensions. 
\setlength{\textfloatsep}{10pt}
\begin{table}[t] \small
\caption{Profiling of GCN Sparse Operations.}
\vspace{-5pt}
\centering
\scalebox{0.9}{
\begin{tabular}{l | r | r | r | r}
\specialrule{.1em}{.05em}{.05em} 
\textbf{Dataset} & \textbf{Aggr. (\%)} & \textbf{Update (\%)} & \textbf{Cache(\%)} & \textbf{Occ.(\%)} 
\\ 
\hline\hline
\multirow{1}{*}{Cora}     &  88.56 &  11.44 & 37.22 & 15.06 \\%
\multirow{1}{*}{Citeseer} &   86.52 & 13.47 & 38.18 & 15.19 
\\ %
\multirow{1}{*}{Pubmed}     &   94.39 & 5.55  & 37.22 & 16.24 \\ 
\specialrule{.1em}{.05em}{.05em} 
\end{tabular}}
\label{tbl: Kernel Profiling of GCN and GAT Sparse Operations}
\vspace{5pt}
\end{table}

From Table~\ref{tbl: Kernel Profiling of GCN and GAT Sparse Operations}, we have several observations:
\hlp{First}, the aggregation phase usually dominates the overall execution of the GNN execution. From these three commonly used GNN datasets, we can see that the aggregation phase usually takes more than 80\% of the overall execution time, which demonstrates the key performance bottleneck of the GNNs is to improve the performance of the sparse neighbor aggregation.
\hlp{Second}, sparse operations in GNNs show very low memory performance. 
The column \textit{Cache} of Table~\ref{tbl: Kernel Profiling of GCN and GAT Sparse Operations} shows 
GNN sparse operations could not well benefit from the GPU cache system, thus, showing a low cache-hit ratio (around 37\%) and frequent global memory access.
\hlp{Third}, sparse operations of GNNs show very inefficient computation.
As described in the column \textit{Occupancy} of Table~\ref{tbl: Kernel Profiling of GCN and GAT Sparse Operations}, the sparse operation of GNNs could hardly keep the GPU busy because 
1) its low computation intensity (the number of non-zero elements in the sparse matrix is generally small);
2) its highly irregular memory access for fetching rows of the dense matrix during the computation, resulting in memory-bound computation;
3) it currently can only leverage CUDA cores for computation, which naturally has limited throughput performance. 
On the other side, this study also points out several potential directions for improving the SpMM performance on GPUs, such as improving the computation intensity (\textit{e.g.}, assigning more workload to each thread/warp/block), boosting memory access efficiency (\textit{e.g.}, crafting specialized memory layout for coalesced memory access), and breaking the computation performance ceiling (\textit{e.g.}, using TCUs). 

\vspace{10pt}
\subsection{Dense GEMM on CUDA Cores/TCUs} 
\label{sect: Dense GEMM on CUDA Cores/TCUs}
While the dense GEMM is mainly utilized for dense NN computation (\textit{e.g.}, linear transformation and convolution), it can also be leveraged for GNN aggregation under some circumstances. For example, when an input graph has a very limited number of nodes, we can directly use the dense adjacency matrix of the graph and accelerate the intrinsically sparse neighbor aggregation computation on CUDA cores/TCUs by calling cuBLAS~\cite{cublas}. However, such an assumption may not hold even for medium-size graphs in real-world GNNs.
\begin{table}[t] \small
\centering
\caption{Medium-size Graphs in GNNs.}
\vspace{-5pt}
\scalebox{0.9}{
\begin{tabular}{l| r | r | r | r}
\specialrule{.1em}{.05em}{.05em} 
\textbf{Dataset} & \textbf{\# Nodes}&  \textbf{\# Edges} & \multicolumn{1}{c|}{\textbf{Memory}} & \textbf{Eff.Comp}
\\ \hline\hline
OVCR-8H & 1,890,931  & 3,946,402 &  14302.48 GB & 0.36\% 
\\ 
Yeast & 1,714,644  & 3,636,546 &  11760.02 GB& 0.32\% \\ 
DD & 	334,925	& 1,686,092  & 448.70 GB & 0.03\% 
\\ 
\specialrule{.1em}{.05em}{.05em} 
\end{tabular}}
\label{tbl: Examples: DGEMM in GNN Sparse Computation.}
\end{table}

As shown in Table~\ref{tbl: Examples: DGEMM in GNN Sparse Computation.}, for these selected datasets, the memory consumption of their dense graph adjacent matrix (as a 2D \texttt{float} array) would easily exceed the device memory constraint of today's GPU (less than 100GB).
Even if we assume the dense adjacent matrix can fit into the GPU memory, the extremely low effective computation (the last column of Table~\ref{tbl: Examples: DGEMM in GNN Sparse Computation.}) would also be a major obstacle for us to achieve high performance. 
We measure the effective computation as $\frac{nnz}{N\times N}$, where $nnz$ is the number of the non-zero elements (indicating edges) in the graph adjacent matrix and $N$ is the number of nodes in the graph. 
The number of $nnz$ is tiny in comparison with the $N\times N$. Therefore, computation and memory access on zero elements are wasted.

\subsection{Hybrid Sparse-Dense Solution}
\label{sect: Hybrid Sparse-Dense Solution}
Another type of work~\cite{tiled-spmm, blocked_spmm} takes the path of mixing the \textit{sparse control} (tile-based iteration) with \textit{Dense GEMM computation}. They first apply a convolution-like (2D sliding window) operation on the adjacent matrix and traverse all possible dense tiles that contain non-zero elements. 
Then, for all identified non-zero tiles, they invoke GEMM on CUDA cores/TCUs for computation. However, this strategy has two shortcomings.
\hlp{First}, the sparse control itself would cause a high overhead.
Based on our empirical study, the non-zero elements are highly scattered on the adjacent matrix of a sparse graph. Therefore, traversing all blocks in a super large adjacent matrix would be time-consuming.
\hlp{Second}, the identified sparse tiles would still waste lots of computation. 
The irregular edge connections of the real-world graphs could hardly fit into these fixed-shape tile frames. 
Therefore, most of the dense tiles would still have very few non-zero elements. 

Inspired by the above studies, we make several design choices in order to achieve high-performance sparse GNN operations.
\hlp{First}, we choose the hybrid sparse-dense solution as the starting point. 
This can give us more flexibility for optimizations at the sparse control (\textit{e.g.}, traversing fewer tiles) and dense computation (\textit{e.g.}, increasing the effective computation/memory access when processing each tile). 
\hlp{Second}, we employ shared memory as the key space for GPU kernel-level data management. 
It can help us to re-organize the irregular GNN input data in a more ``regularized'' way such that both the memory access efficiency and computing performance can be well improved.
\hlp{Third}, we choose TCUs as our major computing unit since they can bring significantly higher computing throughput performance in comparison with CUDA cores. This also indicates the great potential of using TCUs for harvesting more performance gains.
\begin{table}[t] \small
    \caption{Comparison among Sparse GEMM, Dense GEMM, Hybrid Sparse-Dense, and \Mname. Note that \textbf{MC}: Memory Consumption, \textbf{EM}: Effective Memory Access, \textbf{CI}: Computation Intensity, \textbf{EC}: Effective Computation.}
    \centering
    \scalebox{1}{
    \begin{tabular}{l|c c c c}
    \specialrule{.1em}{.05em}{.05em} 
        \textbf{Solution} & \textbf{MC} & \textbf{EM} & \textbf{CI} & \textbf{EC} \\
    \hline\hline
    Sparse GEMM ($\S$\ref{sect: SpMM on CUDA cores}) 
    & Low & Low & Low  & High   \\
    Dense GEMM ($\S$\ref{sect: Dense GEMM on CUDA Cores/TCUs}) 
    & High & High & High & Low  \\
    Hybrid Sparse-Dense ($\S$\ref{sect: Hybrid Sparse-Dense Solution})   
    & High & Low & Low  & High   \\
    \textbf{\Mname}~(This work) & \textbf{Low} & \textbf{High} & \textbf{High} & \textbf{High} \\
    \specialrule{.1em}{.05em}{.05em} 
    \end{tabular}}
    \label{table: comparison with DGEMM and SpMM.}
\end{table}

Finally, we crystallize all of our ideas and insights into \Mname~that effectively coordinates the execution of GNN sparse operations on dense TCU. We show a brief qualitative comparison among \Mname~and the above three solutions in Table~\ref{table: comparison with DGEMM and SpMM.}. 
Note that \textit{Memory Consumption} is the size of memory used by the sparse/dense graph adjacency matrix; The \textit{Effective Memory Access} is the ratio between the size of the accessed data that is actually involved in the later computation and the total data being accessed; The \textit{Computation Intensity} is the ratio of computing operations versus the data being accessed; The \textit{Effective Computation} is the operations for generating the final result versus the total operations.

\section{\Mname~Design}
In this section, we will first give an overview of \Mname~through its high-level programming interface and then detail the TCU-aware GNN algorithm design. 
\begin{figure}[t] \small
\centering
\begin{lstlisting}[caption=Example of a 2-layer GCN in TC-GNN., escapeinside={(*@}{@*)}, style=tt4, label={code: GCN code example.}]
import TCGNN, torch
# include other packages ...
class GCN(torch.nn.Module):
    def __init__(self, inDim, hiDim, outDim):
        self.layer1 = TCGNN.GCNConv(inDim, hiDim)
        self.layer2 = TCGNN.GCNConv(hiDim, outDim)
        self.softmax = torch.nn.Softmax()
        
    def forward(self, tiledGraph, param):
        tiled_adj, X = tiledGraph.adj, tiledGraph.X
        X = self.layer1(X, tiledAdj, param)
        X = self.ReLU(X)
        X = self.layer2(X, tiledAdj, param)
        X = self.softmax(X)
        return X
# Define a two-layer GCN model in TC-GNN.
model = GCN(inDim=100, hiDim=16, outDim=10)
# Load graph and extract input information.
rawGraph, info = TCGNN.Loader(graphFilePath)
# Generate TCU tile and runtime configuration.
tiledGraph, config = TCGNN.Preprocessor(rawGraph, info)
# Run model through forward computation.
predict_y = model(tiledGraph, config)
# Compute loss and accuracy.
# Gradient backpropagation for training.
\end{lstlisting} 
\vspace{-10pt}
\end{figure}
As detailed in Listing~\ref{code: GCN code example.}, \Mname~consists of several key components to facilitate the programming of GNN models on GPU TCUs.
\Mname~introduces a set of pre-built popular GNN layers (e.g., \texttt{TCGNN.GCNConv}) that can be easily connected with some other existing neural network layers (e.g., \texttt{ReLU} and \texttt{softmax}), to help users define their own GNN model quickly. For those non-conventional GNN layers, users can directly use our low-level APIs (e.g., \texttt{TCGNN.spmm} and \texttt{TCGNN.sddmm}) to express the GNN computation easily.
\Mname~introduces an input \textbf{\code{Loader}} to load the GNN input graph as a \textit{rawGraph} and capture the key input information for system-level optimizations.
\Mname~incorporates a \textbf{\code{Preprocessor}} to build tiles from \textit{rawGraph} and generate TCU-aware \textit{tiledGraph} ($\S$\ref{sect: TCU-Aware Sparse Graph Translation}), and optimize runtime configuration (e.g., warps per block) for our TCU-tailored GPU kernel ($\S$\ref{sect: TCU-tailored GNN Computation} and $\S$\ref{sect: TCU-centric Workload Mapping}) based on input.
Finally, we train the initialized GNN model defined in \Mname~as the regular GNN models defined in other frameworks through forward and backward computation.
\begin{algorithm}[t] 
  \caption{TCU-aware Sparse Graph Translation.} \small
  \label{algo: Sparse Graph Translation.}
\SetAlgoLined
  \SetKwInOut{Input}{input}
  \SetKwInOut{Output}{output}
  \Input{Graph adjacent matrix $\mathbf{A}$ ($\mathit{nodePointer}$, $\mathit{edgeList}$).}
  \Output{Result of $\mathit{winPartition}$ and $\mathit{edgeToCol}$.}
    \tcc{Compute the total number of row windows.}
    $\mathit{numRowWin}$ = \textbf{ceil}($\mathit{numNodes/winSize}$)\;
    \For{$winId$ \textbf{in} $numRowWin$}{
        \tcc{EdgeIndex range of the current rowWindow.}
        $winStart$ = $nodePointer[winId*winSize]$\;
        $winEnd$ = $nodePointer[(winId+1)*winSize]$\;
    
        \tcc{Sort the edges of the current rowWindow.}
        $eArray$ = \textbf{Sort}($winStart$, $winEnd$, $edgeList$)\;
        \tcc{Deduplicate edges of the current rowWindow.}
        $eArrClean$ = \textbf{Deduplication}($eArray$)\;
        
        \tcc{\#TC blocks in the current rowWindow.}
        $winPartition[winId]$ = \textbf{ceil}($eArrClean.size/TC\_BLK\_w$)\;
                
        \tcc{Edges-to-columnID mapping in TC Blocks.}
        \For{$eIndex$ \text{\textbf{in}} [$\mathit{winStart}$, $\mathit{winEnd}$]}{
            $eid$ = $edgeList[eIndex]$\;
            $edgeToCol[eIndex]$ = $eArrClean[eid]$;
        }
    }
\end{algorithm}
\begin{figure*}[t] \small
    \centering
    \includegraphics[width=2\columnwidth]{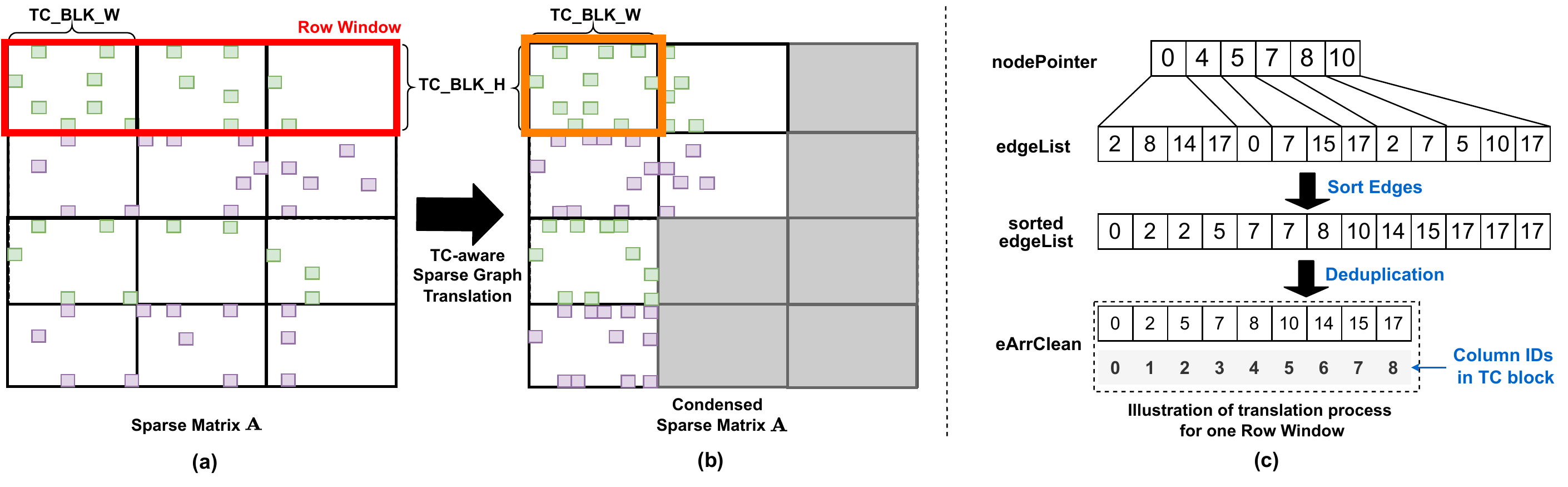}
    \vspace{-10pt}
    \caption{Illustration of Sparse Graph Translation. Note that the grey area indicates the TCU blocks that will be directly skipped.}
    \label{fig: Graph Translation Example.}
    \vspace{-4pt}
\end{figure*}

\begin{algorithm}[t]  \footnotesize
  \caption{TC-GNN Neighbor Aggregation.}
  \label{algo: Neighbor Aggregation on Translated Graph.}
\SetAlgoLined
  \SetKwInOut{Input}{input}
  \SetKwInOut{Output}{output}
  \Input{Condensed graph structure ($nodePointer$, $edgeList$, $edgeToCol$, $winPartition$) and node embedding matrix ($\mathbf{X}$).}
  \Output{Updated node embedding matrix ($\mathbf{\hat{X}}$).}
    \tcc{Traverse through all row windows.}
    \For{$winId$ \textbf{in} $numRowWindows$}{
        \tcc{\#TC blocks of the row window.}
         $numTCblocks$ = $winPartition[winId]$ \;
        \tcc{Edge range of TC blocks of the row window.}
         $edgeRan$ = \textbf{GetEdgeRange}($nodePointer$, $winId$)\;
        \For{$TCblkId$ \textbf{in} $numTCblocks$}{
            \tcc{The edgeList chunk in current TC block.}
            $edgeChunk$ = \textbf{GetChunk}($edgeList$, $edgeRan$, $TCblkId$)\;
            \tcc{Neighbor node Ids in current TC block.}
            $colToNId$ = \textbf{GetNeighbors}($edgeChunk$, $edgeToCol$)\;
            \tcc{Initiate a dense tile ($ATile$).}
            $ATile$ = \textbf{InitSparse}($edgeChunk$, $winId$)\;
            \tcc{Initiate a dense tile ($XTile$).}
            $XTile$, $colId$ = \textbf{FetchDense}($colToNId$, $X$)\;
            \tcc{Compute $XnewTile$ via TCU GEMM.}
            $XnewTile$ = \textbf{TCcompute}($ATile$, $XTile$)\;
            \tcc{Store $XnewTile$ of $\hat{X}$.}
            $\hat{X}$ = \textbf{StoreDense}($XNewTile$, $winId$, $colId$)\;
        }
    }
\end{algorithm}
\subsection{TCU-aware Sparse Graph Translation}
\label{sect: TCU-Aware Sparse Graph Translation}
As the major component of \Mname, we introduce a novel \textit{Sparse Graph Translation} (SGT) technique to facilitate the TCU acceleration of GNNs. 
Our core idea is that \textit{the pattern of the graph sparsity can be well-tuned for TCU computation through effective graph structural manipulation meanwhile guaranteeing output correctness}. 
{Our key observation is that neighbor sharing is common in real-world graphs and has been exploited for various tasks like link prediction~\cite{zhang2018link}. Our evaluated datasets (Section~\ref{sect: Evaluation}) have 18\% to 47\% (averaged 29\%) neighbor similarity.}
Specifically, we condense (remap) the highly-scattered neighbor ids into highly-condensed new neighbor ids that can facilitate the dense TCU computation paradigm. Also, such condensing should not compromise any original information (\textit{e.g.}, edge connections) and can generate the exact output as the conventional design. 

As exemplified in Figure~\ref{fig: Graph Translation Example.}\subfig{a} and Figure~\ref{fig: Graph Translation Example.}\subfig{b}, we take the regular graph in CSR format as the input and condense the columns of each row window (in the red-colored rectangular box) to build TCU blocks ($TC\_block$) (\textit{a.k.a.}, the input operand shape of a single MMA instruction), in the orange-colored rectangular box. 
$\mathit{nodePointer}$ is the row pointer array $\mathit{edgeList}$ is the edges of each node stored continuously.
In this paper, we demonstrate the use of standard MMA shape for TF-32 of TCU on Ampere GPU architecture, and other MMA shapes~\cite{TC-rules} can also be used under different precision (\textit{e.g.}, \texttt{half} and \texttt{int8}) and GPU architecture (\textit{e.g.}, Turing).  

SGT takes several steps for processing each row window, as detailed in Algorithm~\ref{algo: Sparse Graph Translation.} and visualized in Figure~\ref{fig: Graph Translation Example.}\subfig{c}. \textit{winPartition} is an array for maintaining the number of TC blocks in each row window. \textit{edgeToCol} is an array for maintaining the mapping between the edges and their corresponding position in the graph after SGT. {Note that \textit{edgeToCol} has the same length as \textit{edgeList} but with \textit{column-id} from \textit{eArrClean}.
\textit{colToRow} maps \textit{column-id} of adjacency matrices to the \textit{row-id} of embedding matrices.
}
We choose the size of the row window ($winSize$=\textit{TC\_BLK\_H}) and column width (\textit{TC\_BLK\_W}) according to TCU MMA specification (\textit{e.g.}, \textit{TC\_BLK\_H}=16, \textit{TC\_BLK\_W}=8 in TF-32). 
After condensing the graph within each row window, the time complexity of sliding the $TC\_block$ can be reduced from $\mathcal{O}(\frac{N}{TC\_BLK\_W})$ to only $\mathcal{O}(\frac{nnz_{unique}}{TC\_BLK\_W})$, where $N$ is the total number of nodes in the graph and $nnz_{unique}$ is the size of the unique neighbor within the current row window, which equals $eArrClean.size$ in Algorithm~\ref{algo: Sparse Graph Translation.}. 
The density (computation intensity) of each identified TCU block can be largely improved. Considering the case in Figure~\ref{fig: Graph Translation Example.}, after the sparse graph translation, we can achieve $2\times$ higher density on individual TCU blocks (Figure~\ref{fig: Graph Translation Example.}\subfig{b}) compared with the original one (Figure~\ref{fig: Graph Translation Example.}\subfig{a}). 

Compared to existing sparse matrix formats (e.g., Blocked-Ellpack~\cite{blocked_spmm}) which use the regular matrix tiles to cover the irregularly scattered non-zero elements,   SGT reduces the irregularity of non-zero-elements layout to fit them into fewer number TCU blocks, thus, reducing the unnecessary computation and memory overhead.
SGT is applicable for both the SpMM and SDDMM in GNN sparse operations and can be easily parallelized because the processing of individual row windows is independent. 
In most cases, SGT only needs to execute once and its result can be reused across many epochs/rounds of GNN training/inference.

{Additionally, SGT can be generally used with other accelerators (e.g., AMD-GPUs with matrixCore and TPUs) that offer similar dense MM primitives.
CPUs have no direct alternative to TensorCore-like MM primitives. However, with AVX-vectorized instructions, CPUs can benefit from SGT by setting \textit{BLK\_H}=1 and \textit{BLK\_W}=(\#elements-per-AVX-instruction).
TC-GNN currently targets GNN training. SGT is conducted once before training. SGT cost can be offset by training iterations (averaged 2\% for 200 iterations as DGL).}

\subsection{TCU-tailored GNN Computation}
\label{sect: TCU-tailored GNN Computation}
Besides the effective way to condense the sparse tiles, the next major challenge is \textit{how to tailor the computation schedule of GNN algorithms} so that we can capitalize on the performance of  condensed sparse graphs and the powerful TCUs. We focus on two major types of computation in GNNs.

\textbf{Neighbor Aggregation} The major part of GNN sparse computing is neighbor aggregation, which can generally be formalized as SpMM operations by many state-of-the-art frameworks~\cite{wang2019dgl}. And they employ the cuSPARSE~\cite{cusparse} on CUDA cores as a black-box technique for supporting sparse GNN computation. 
In contrast, our \Mname~design targets at TCU for the major neighbor aggregation computation which demands a specialized algorithmic design. \Mname~focuses on maximizing the net performance gains by gracefully batching the originally highly irregular SpMM as dense GEMM computation and solving it on TCU effectively.
As illustrated in Algorithm~\ref{algo: Neighbor Aggregation on Translated Graph.}, the node aggregation processes all TC blocks from each row window. \textit{nodePointer} and \textit{edgeList} are directly from graph CSR, while \textit{edgeToCol} and \textit{winPartition} are generated from SGT discussed in the previous section.
Note that $\textbf{InitSparse}$ is to initialize a sparse tile in dense format according to the translated graph structure of the current TC block. 
Meanwhile, $\textbf{FetchDense}$ returns a dense node embedding matrix tile $XTile$ for TCU computation, and the corresponding column range $colId$ (embedding dimension range) of matrix $\mathbf{X}$. This is to handle the case that the width of one $XTile$ could not cover the full-width (all dimensions) of $\mathbf{X}$. Therefore, the $colId$ will be used to put the current TCU computation output to the correct location in the updated embedding matrix $\mathbf{\hat{X}}$.
 \begin{algorithm}[t] \footnotesize
  \caption{TC-GNN Edge Feature Computation.} 
  \label{algo: Neighbor Feature Computing on Translated Graph.}
\SetAlgoLined
  \SetKwInOut{Input}{input}
  \SetKwInOut{Output}{output}
  \Input{Condensed graph data ($\mathit{nodePointer}$, $\mathit{edgeList}$, $\mathit{edgeToCol}$, $\mathit{winPartition}$) and node embedding matrix ($\mathbf{X}$).}
  \Output{Edge Feature List ($\mathit{edgeValList}$).}
    \tcc{Traverse through all row windows.}
    \For{$winId$ \textbf{in} $numRowWin$}{
        \tcc{\#TC blocks in the row window.}
         $numTCblocks$ = $winPartition[winId]$ \;
        \tcc{Edge range of TC blocks of the row window.}
         $edgeRan$ = \textbf{GetEdgeRange}($nodePointer$, $winId$)\;
        \For{$TCblkId$ \textbf{in} $numTCblocks$}{
            \tcc{EdgeList chunk in current TC block.}
            $edgeChunk$ = \textbf{GetChunk}($edgeList$, $edgeRan$, $TCblkId$)\;
            \tcc{Neighbor node Ids in current TC block.}
            $colToNId$ = \textbf{GetNeighbors}($edgeChunk$, $edgeToCol$)\;
            \tcc{Fetch a dense tile ($XTile_A$).}
            $XTile_A$ = \textbf{FetchDenseRow}($winId$, $TCblkId$, $X$)\;
            \tcc{Fetch a dense tile ($XTile_B$).}
            $XTile_B$ = \textbf{FetchDenseCol}($colToNId$, $edgeToCol$, $X$)\;
            \tcc{Compute $edgeValTile$ via TCU GEMM.}
            $edgeValTile$ = \textbf{TCcompute}($XTile_A$, $XTile_B$)\;
            \tcc{Store $edgeValTile$ to $edgeValList$.}
            \textbf{StoreSparse}($edgeValList$, $edgeValTile$, \\ 
                            \hspace{45pt}    $edgeList$, $edgeToCol$)\;
        }
    }
\end{algorithm}

\textbf{Edge Feature Computing} 
Previous research~\cite{AGNN,GATConv} has demonstrated the great importance of incorporating the edge feature for a better GNN model algorithmic performance (\textit{e.g.}, accuracy, and F1-score). 
The underlying building block to generate edge features is the Sampled Dense-Dense Matrix Multiplication (SDDMM)-like operation. 
In \Mname, we support SDDMM with the collaboration of the above sparse graph translation and TCU-tailored algorithm design, as described in Algorithm~\ref{algo: Neighbor Feature Computing on Translated Graph.}. 
The overall algorithm structure and inputs are similar to the above neighbor aggregation. The major difference is the output. In the case of neighbor aggregation, our output is the updated dense node embedding matrix ($\mathbf{\hat{X}}$), where edge feature computing will generate a sparse output with the same shape as the graph edge lists.
Note that fetching the $XTile_A$ only needs to consecutively access the node embedding matrix $\mathbf{A}$ by rows while fetching the $XTile_B$ requires first computing the TCU block column-id to node-id ($\mathit{colToNId}$) to fetch the corresponding neighbor node embeddings from the same node embedding matrix $\mathbf{X}$. 

{Despite the dataflow similarity with dense-GEMM computation (e.g., CUTLASS~\cite{cutlass}), \Mname~has to overcome the limited parallelism (imbalance workload) and sparse/irregular access with novel algorithmic and kernel designs. While these challenges are absent in dense-GEMM computation with naturally high parallelism and data-access locality.} 
\subsection{TCU-centric Workload Mapping}
\label{sect: TCU-centric Workload Mapping}
In collaborating with our TCU-tailored algorithm design, an effective mapping of our algorithmic design to low-level GPU primitives is indispensable for high-performance delivery. We discuss two key techniques: \textit{GPU-aware Workload Decomposition} and \textit{TCU-optimized dataflow design}.
\begin{figure*}[t] \small
    \centering
    \includegraphics[width=2\columnwidth]{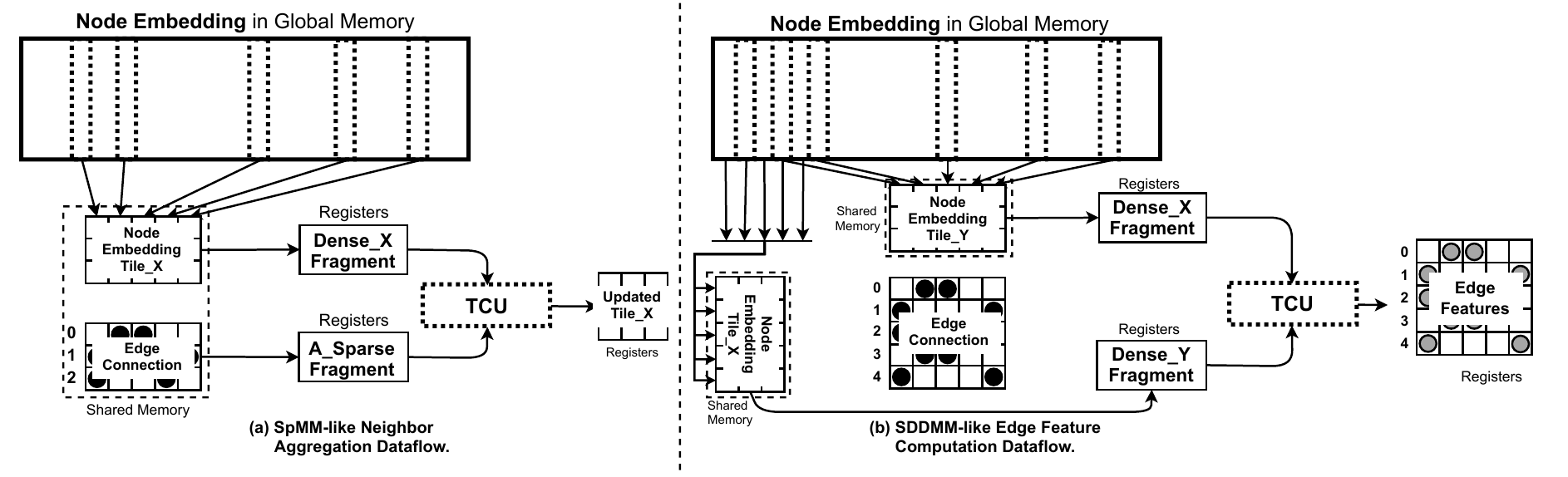}
    \vspace{-5pt}
    \caption{TCU-optimized Dataflow Design for (a) Neighbor Aggregation and (b) Edge Feature Computing in GNNs.}
    \label{fig: Dataflow Design in Sparse GNN Operations.}
    \vspace{-5pt}
\end{figure*}

\subsubsection{GPU-aware Workload Decomposition}
Different from previous work~\cite{wang2019dgl, pyG} focusing on CUDA cores only, \Mname~highlights itself with CUDA core and TCU collaboration through effective two-level workload mapping. 
The idea is based on the fact that CUDA cores work in SIMT fashion and are operated by individual threads, while TCU designated for GEMM computation requires collaboration from a warp of threads (32 threads). 
Our key design principle is to \textit{mix these two types of computing units as a single GPU kernel}, which can efficiently coordinate the kernel execution at different levels of execution granularity.

In \Mname, we operate CUDA cores by thread blocks and manage TCU by thread warps. 
Specifically, threads running CUDA cores from the same thread block will load data (\textit{e.g.}, edges) from the global memory to shared memory. Note that in our design we assign each row window (discussed in $\S$\ref{sect: TCU-Aware Sparse Graph Translation}) to one thread block. The number of threads in each block should be divisible by the number of threads in each warp (32) for better performance.
Once threads running on CUDA cores (CUDA-core threads) finish the data loading, threads from each warp (TCU threads) will operate TCU for GEMM computation (including loading the data from the shared memory to thread-local registers (fragments), applying GEMM computation on data in registers, accumulating results on registers, and storing the final results back to global memory). Note that there would be a large overlap of the CUDA-core threads and TCU threads, both of which are threads from the same blocks but running at a different time frames. In general, we use more CUDA-core threads than TCU threads considering that global memory access demands more parallelization.

There are two major benefits of such two-level workload decomposition. 
First, threads from the same block can work together to improve the memory access parallelization to better utilize memory bandwidth. 
Second, warps from the same block can reuse the loaded data, including the information (\textit{e.g.}, column index mapping) of the translated graph and the tiles from the dense node embedding matrix. Therefore, we can avoid redundant high-cost global memory operations.

\subsubsection{TCU-optimized Dataflow Design}
As the major technique to improve the GPU performance, shared memory is customized for our TCU-based sparse kernel design for re-organizing data layout for dense TCU computation and reducing the redundant global memory traffic. 
Our design takes the TCU specialty into careful consideration from two aspects, 
1) the input matrix tile size of the TCU, which is M(16)$\times$N(16)$\times$K(8) in the case of TF-32, and 
2) the tile fragment layout for fast computation. 
The common practice of the loaded tile A and B are stored in row-major and column-major for better performance. 
Next, we will detail our TCU-optimized dataflow design for both neighbor aggregation and edge feature computation.

\textbf{Neighbor Aggregation}  
In Figure~\ref{fig: Dataflow Design in Sparse GNN Operations.}\subfig{a}, shared memory is mainly used for caching several most frequently used information, including the tile of sparse matrix A (\texttt{sparse\_A}), the column-id of the sparse matrix $\mathbf{A}$ to row-id of node embedding matrix $\mathbf{X}$ (\texttt{sparse\_AToX\_index}), and the dense tile of $\mathbf{X}$ (\texttt{dense\_X}). 
When handling each TCU block, we assign all threads from the same block of threads for loading the sparse tile while allowing several warps to concurrently load the dense row tile from the matrix $X$. 
The reasons for enforcing such caching are two-fold. 
First, it can bridge the gap between the sparse graph data and the dense GEMM computing that requires continuous data layout. 
For example, the adjacent matrix $\mathbf{A}$ is input as CSR format that \textbf{cannot} be directed feed to TCU GEMM computation, therefore, we use a shared memory \texttt{sparse\_A} to initialize its equivalent dense tile. Similarly, we cache rows of $\mathbf{X}$ according to the columns of  $\mathbf{A}$ to the row of $\mathbf{X}$ mapping after our sparse graph translation, where originally scattered columns of $\mathbf{A}$ (the rows of $\mathbf{X}$) are condensed.
Second, it can enable data reuse on \texttt{sparse\_AToX\_index} and \texttt{sparse\_A}. This is because in general, the \texttt{BLK\_H} (16) cannot cover all dimensions of a node embedding (\textit{e.g.}, 64),
multiple warps will be initiated of the same block to operate TCU in parallel to work on non-overlapped dense tiles while sharing the same sparse adjacency matrix tile.
\textbf{Edge Feature Computation}  
Similar to the shared memory design in neighbor aggregation, for edge feature computing, as visualized in Figure~\ref{fig: Dataflow Design in Sparse GNN Operations.}\subfig{b}, the shared memory is utilized for \texttt{sparse\_A}, \texttt{sparse\_AToX\_index}, and \texttt{dense\_X}. We assign all threads from the same block of threads for loading the sparse tile while allowing several warps to concurrently load the dense row tile from the matrix $\mathbf{X}$. Compared with dataflow design in neighbor aggregation, edge feature computing demonstrates several differences. 

\hlp{First}, the sizes of \texttt{sparse\_A} are different. 
In the neighbor aggregation computation, the sparse matrix $\mathbf{A}$ is used as one operand in the SpMM-like computation, therefore, the minimal processing granularity is $16\times 8$, while in edge feature computing by following SDDMM-like operation, the sparse matrix $\mathbf{A}$ serves as the output matrix, thus, maintaining the minimum processing granularity is $16\times 16$. To reuse the same translated sparse graph as SpMM, we need to recalculate the total number of TC blocks.
\hlp{Second}, iterations along the embedding dimension would be different. Compared with neighbor aggregation, edge feature computing requires the result accumulation along the embedding dimension. 
The result will only be output until all iterations have finished. In neighbor aggregation, the node embedding vector is divided among several warps, each of which will output their aggregation result to non-overlapped embedding dimension ranges in parallel.
\hlp{Third}, the output format has changed. Compared with SpMM-like neighbor aggregation which directly output computing results as an updated dense matrix $\mathbf{\hat{X}}$, SDDMM-like edge feature computing requires a sparse format (the same shape as \texttt{edgeList}) output for compatibility with neighbor aggregation and memory space. Therefore, one more step of dense-to-sparse translation is employed.
\begin{table}[t] \small	
\vspace{-5pt}
\caption{Datasets for evaluation.}
\vspace{-5pt}
\centering
\scalebox{0.79}{
 \begin{tabular}{ c | l c r r r r}
\specialrule{.1em}{.05em}{.05em} 
\textbf{Type} & \textbf{Dataset} & \textbf{Abbr.} & \textbf{\#Vertex} & \textbf{\#Edge} & \textbf{Dim.} & \textbf{{\#Class}}\\
\hline
\multirow{4}{*}{\textbf{I}} & Citeseer    & CR & 3,327	    & 9,464	    & 3703 & 6      \\
& Cora & CO	    & 2,708     & 10,858	& 1433 & 7      \\
& Pubmed & PB	    & 19,717	& 88,676	& 500  & 3      \\
& PPI & PI	        & 56,944	& 818,716	& 50   & 121    \\
\hline
\hline

\multirow{5}{*}{\textbf{II}} 
& PROTEINS\_full & PR	&   43,471	       & 162,088	&   29	    & 2 \\
& OVCAR-8H & OV	    &   1,890,931	   & 3,946,402	&   66	    & 2 \\
& Yeast & YT        &   1,714,644	   & 3,636,546	&   74	    & 2 \\
& DD	& DD            &   334,925	       & 1,686,092	&   89	    & 2 \\
& YeastH & YH	        &   3,139,988      & 6,487,230	&   75	    & 2 \\
\hline
\hline

\multirow{5}{*}{\textbf{III}} 
& amazon0505 & AZ	    & 410,236	& 4,878,875	    & 96  & 22 \\
& artist& AT	        & 50,515	& 1,638,396	    & 100 & 12 \\
& com-amazon & CA	    & 334,863	& 1,851,744	    & 96  & 22 \\
& soc-BlogCatalog & SC	& 88,784	& 2,093,195	    & 128 & 39 \\
& amazon0601	& AO    & 403,394	& 3,387,388	    & 96 & 22 \\
\specialrule{.1em}{.05em}{.05em} 
\end{tabular}}
\label{table: Evaluation Dataset}
\vspace{8pt}
\end{table}

\section{Evaluation}
\label{sect: Evaluation}
\textbf{Benchmarks: }
We choose two representative GNN models widely used by previous work~\cite{wang2019dgl,pyG,ma2019neugraph} on \textit{node classification} tasks. 
Specifically, 
{1) Graph Convolutional Network (\textbf{\underline{GCN}})}~\cite{GCNConv} is one of the most popular GNN model architectures. 
It is also the key backbone for many other GNNs (e.g., GraphSAGE~\cite{SageConv} and differentiable pooling (Diffpool)~\cite{diffpool}). 
Therefore, improving the performance of GCN will also benefit a broad range of GNNs. For GCN evaluation, we use the setting: \textit{2 layers with 16 hidden dimensions per layer}, which is also the setting from the original paper~\cite{GCNConv}.
{2) Attention-based Graph Neural Network (\textbf{\underline{AGNN}})}~\cite{AGNN}.
AGNN differs from GCN in its aggregation function, which computes edge features (via embedding vector dot-product between source and destination vertices) before the node aggregation. 
AGNN is also the reference architecture for many other recent GNNs for better model algorithmic performance. 
For AGNN, we use: 
\textit{4 layers with 32 hidden dimensions per layer}.

\textbf{Baselines: } 
1) Deep Graph Library (\underline{\textbf{DGL}})~\cite{wang2019dgl} is the state-of-the-art GNN framework on GPUs, which is built with the high-performance CUDA-core-based cuSPARSE~\cite{cusparse} library as the backend and uses PyTorch~\cite{pytorch} as its front-end programming interface. 
DGL significantly outperforms other existing GNN frameworks~\cite{pyG} over various datasets on many mainstream GNN model architectures. 
Therefore, we make an in-depth comparison with DGL.
2) PyTorch Geometric (\underline{\textbf{PyG}})~\cite{pyG} is another GNN framework. 
PyG leverages torch-scatter~\cite{torch-scatter} library (highly-engineered CUDA-core kernel) as the backend support, which highlights its performance on batched small graph settings;
3) Blocked-SpMM~\cite{blocked_spmm} (\underline{\textbf{bSpMM}}) accelerates SpMM on TCU. It is included in the recent update on the cuSPARSE library. bSpMM requires the sparse matrix with Blocked-Ellpack format for computation. Its computation on non-zero blocks can be seen as the hybrid sparse-dense solution ($\S$\ref{sect: Hybrid Sparse-Dense Solution}). 
Note that the bSpMM has not been incorporated into any existing GNN frameworks. We also compare \Mname~with \underline{\textbf{tSparse}}~\cite{zachariadis2020accelerating} and \underline{\textbf{Triton}}~\cite{Triton} for non-vendor-developed highly optimized kernels on TCUs.

\textbf{Datasets, Platforms, and Metrics: }
We cover three types of datasets (Table~\ref{table: Evaluation Dataset}), which have been used in previous GNN-related work~\cite{wang2019dgl, pyG, ma2019neugraph}.
Specifically, \textbf{\underline{Type I}} graphs are the typical datasets used by previous GNN algorithm papers~\cite{GCNConv, GINConv, SageConv}. 
They are usually small in the number of nodes and edges, but rich in node embedding information with high dimensionality. 
\textbf{\underline{Type II}} graphs~\cite{KKMMN2016} are the popular benchmark datasets for graph kernels and are selected as the built-in datasets for PyG~\cite{pyG}. Each dataset consists of a set of small graphs, which only have intra-graph edge connections without inter-graph edge connections. 
\textbf{\underline{Type III}} graphs~\cite{snapnets,GCNConv} are large in terms of the number of nodes and edges. These graphs demonstrate high irregularity in its structures, which are challenging for most of the existing GNN frameworks. 
The core design of \Mname~consists of around 2.5K lines of code. \Mname~backend is implemented with C++ and CUDA C, and its front end is implemented in Python. 
Our major evaluation platform is a server with an 8-core 16-thread Intel Xeon Silver 4110 CPU and an NVIDIA RTX3090 GPU.
To measure the performance speedup, we calculate the average latency of 200 end-to-end runs.
\begin{figure*}[t] \small
    \centering
    \subfloat[]{\includegraphics[width=0.33\textwidth, trim=0 0cm 0 0 ]{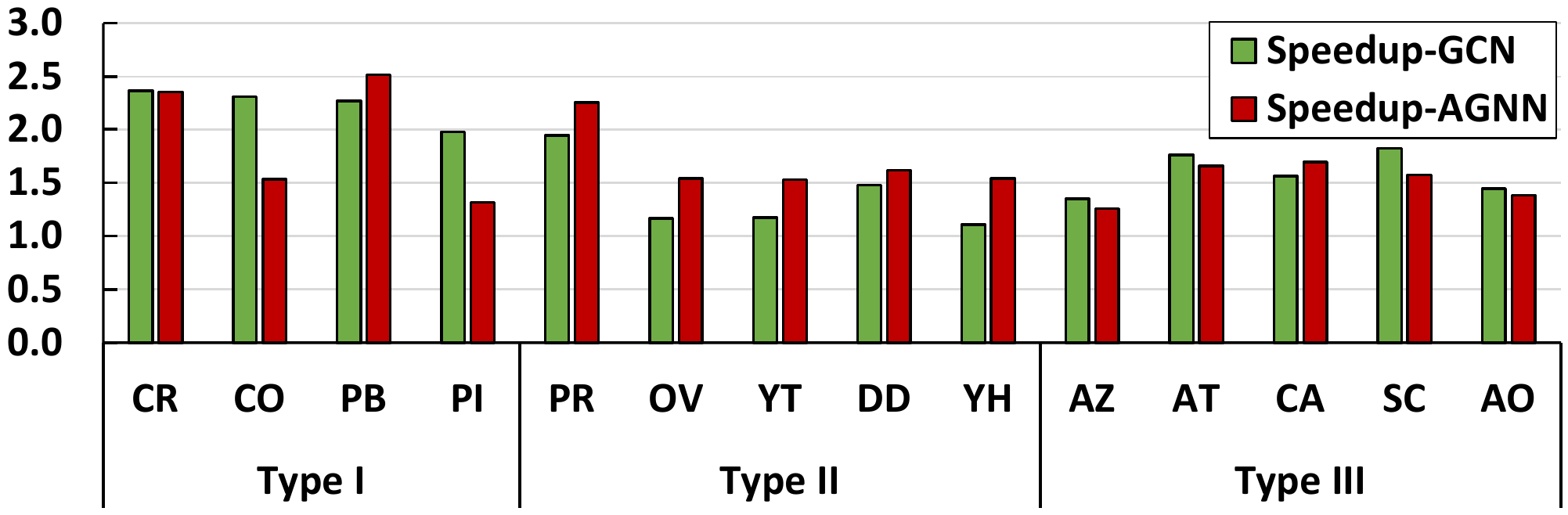}}
    \subfloat[]{\includegraphics[width=0.33\textwidth, trim=0 0cm 0 0cm]{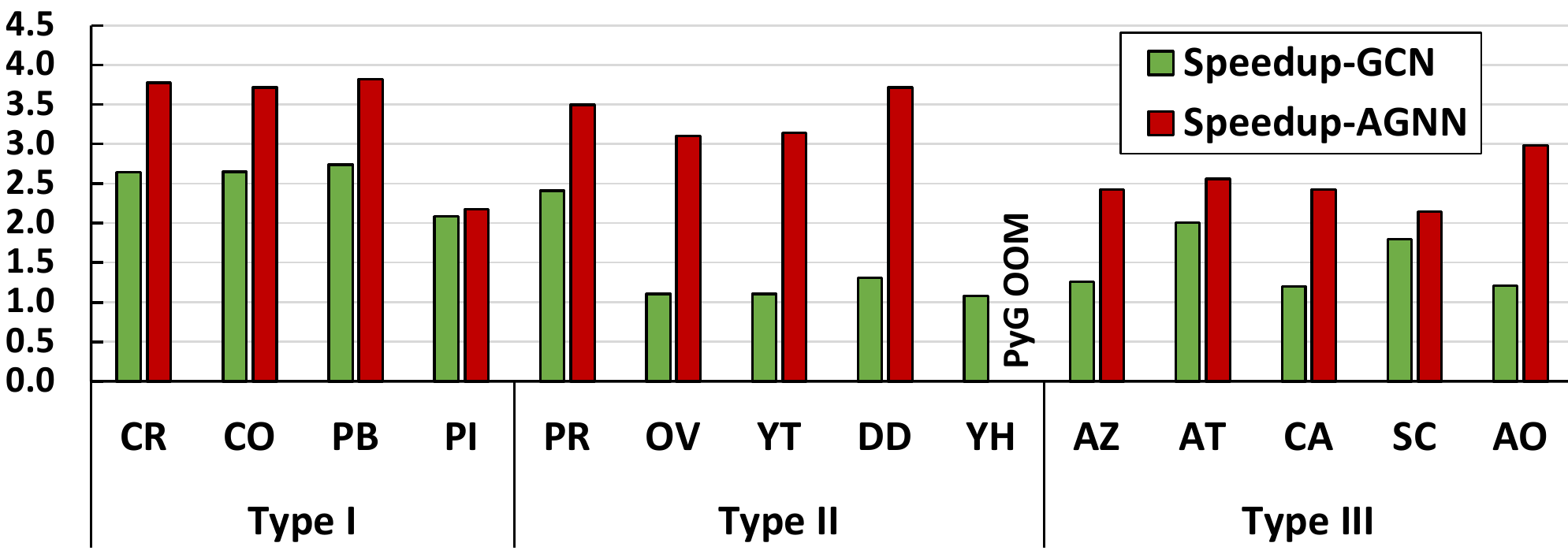}}
    \subfloat[]{\includegraphics[width=0.33\textwidth,trim=0cm 0cm 0 0]{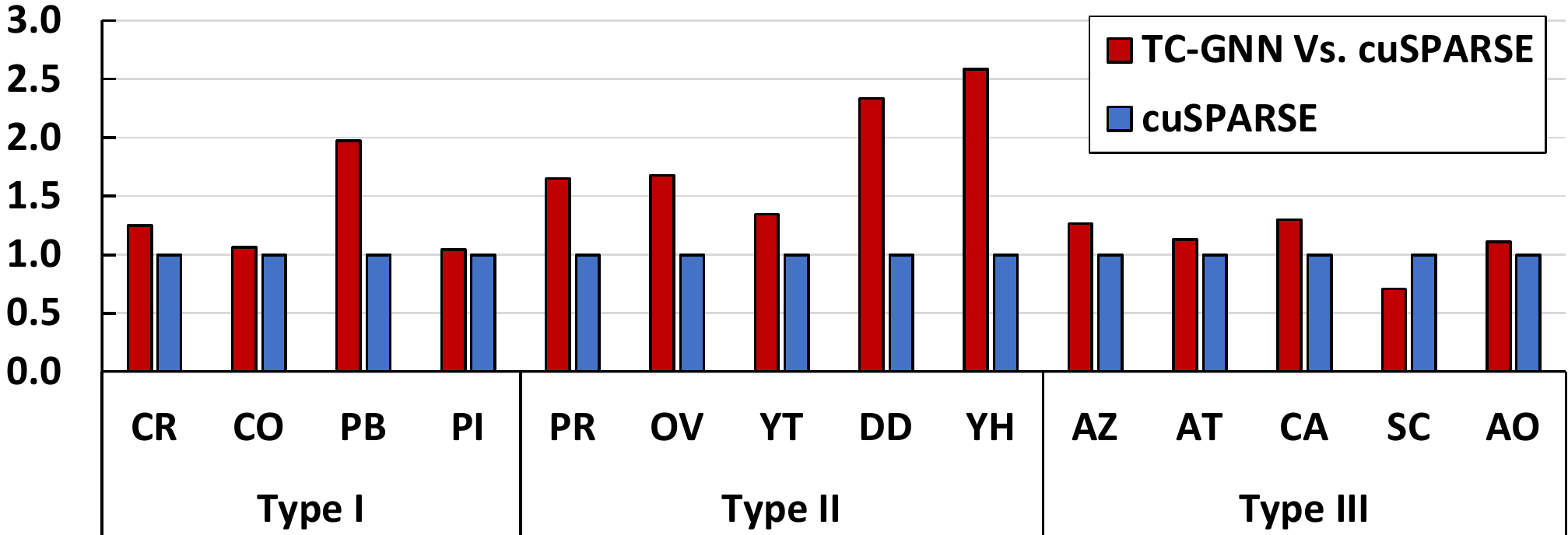}}
    %
    \caption{Speedup over (a) DGL and (b) PyG on GCN and AGNN; (c) Speedup over cuSPARSE bSpMM on TCUs.}
    \label{fig: Performance Comparison.}
    \vspace{5pt}
\end{figure*} 

\subsection{Compared with DGL}
\label{sect: compared with DGL}
Figure~\ref{fig: Performance Comparison.}\subfig{a} shows that \Mname~achieves $1.70\times$ speedup on average compared to DGL over three types of datasets across GCN and AGNN models on end-to-end training. Our kernel profiling via Nsight Compute shows that \Mname~achieves high SM occupancy (averaged 85.28\%), which is on average 21.05\%  higher compared to DGL across all datasets.

\textbf{Type I Graphs:}  
The performance improvements against DGL are significantly higher for GCN (on average $2.23\times$) compared to AGNN (on average $1.93\times$). 
The major reason is their different GNN computation patterns. 
GCN only consists of a neighbor aggregation phase (SpMM-like operation) and a node update phase (GEMM operation).
Whereas in the AGNN, the aggregation phase would also require an additional edge attention value (feature) computation based on SDDMM-like operations. 
Compared with SpMM-like operations, edge attention computation in SDDMM is more sensitive to the irregular sparse graph structure because of much more intensive computations and memory access. Thus, the performance improvement is relatively lower. 

\textbf{Type II Graphs:} 
\Mname~achieves averaged $1.38\times$ speedup on GCN and $1.70\times$ speedup on AGNN for the Type II graphs. 
Speedup on Type II graphs is relatively lower compared with Type I, since Type II datasets consist of a set of small graphs with very dense intra-graph connections but no inter-graph edges. 
This leads to a lower benefit from the sparse graph translation that would show more effectiveness on highly irregular and sparse graphs. 
Such a clustered graph structure would also benefit cuSPARSE due to more efficient memory access, \textit{i.e.}, less irregular data fetching from the sparse matrix.
In addition, for AGNN, \Mname~can still demonstrate evident performance benefits over the DGL (CUDA core only) that can mainly contribute to TCU-based SDDMM-like designs that can fully exploit the power of GPU through an effective TCU and CUDA core collaboration. 

\textbf{Type III Graphs:} 
The speedup is also evident (on average 1.59$\times$ for GCN and average 1.51$\times$ for AGNN) on graphs with a large number of nodes and edges and irregular graph structures. 
The reason is the high overhead global memory access can be well reduced through our spare graph translation. 
Besides, our dimension-split strategy further facilitates efficient workload sharing among warps by improving the data spatial/temporal locality. 
On the dataset AT and SC, which have a higher average degree within Type III datasets, we notice a better speedup performance for both GCN and AGNN. This is because 
1) more neighbors per node can lead to a higher density of non-zero elements within each tile/fragment. Thus, it can fully exploit the computation benefits of each TCU GEMM operation;  
2) it can also facilitate more efficient memory access. For example, in AGNN, fetching one dense embedding $x$ from the dense matrix $X$ can be reused more times by applying a dot-product between $x$ and many columns of the dense matrix $X^{T}$ (neighbors embeddings).

Additionally, our performance breakdown analysis shows that for graphs with highly scattered and irregular edge distribution, such as Type I and III graphs, SGT would contribute more (averaged 64\%) to the overall performance improvements since it helps significantly reduce the unnecessary workload.
For graphs with highly dense and more regular edge connections, such as Type II datasets, SGT contributes relatively minor (averaged 23\%) to the overall performance since it could not squeeze out more redundant computations from already condensed edge tiles.
\begin{table}[t] \small
\centering
\caption{Compare TC-GNN with tSparse and Triton.}
\vspace{-5pt}
\scalebox{0.9}{
\begin{tabular}{l|r r r}
\specialrule{.1em}{.05em}{.05em} 
\textbf{Dataset} &   \textbf{tSparse (ms)} & \textbf{Triton (ms)} & \textbf{TC-GNN (ms)} 
\\ \hline \hline
AZ  & 18.60 & 31.64 & 4.09 \\
AT  & 9.15 & 12.86 & 3.06 \\
CA  & 13.84 & 15.50 & 3.26 \\ 
SC & 9.74 & 14.38 & 3.59 \\ 
AO & 11.93 & 21.78 & 3.41 \\ 
\specialrule{.1em}{.05em}{.05em} 
\end{tabular}}
\label{tbl: Compare with tSparse, Triton}
\end{table}
\subsection{Compared with other baselines}
\label{sect: compared with other baselines}
\textbf{Compared with PyG} Figure~\ref{fig: Performance Comparison.}\subfig{b} shows \Mname~can outperform PyG with an average of $1.76\times$ speedup on GCN and an average of $2.82\times$ speedup on AGNN. 
For GCN, \Mname~achieves significant speedup on datasets with high-dimensional node embedding, such as \textit{Yeast (YT)}, through effective TCU acceleration through a TCU-aware sparse graph translation while reducing the synchronization overhead by employing our highly parallelized TCU-tailored algorithm design. 
PyG, however, achieves inferior performance because its underlying GPU kernel can only leverage CUDA cores, thus, intrinsically bounded by CUDA core performance. 

\textbf{Compared with cuSPARSE bSpMM}
%
Figure~\ref{fig: Performance Comparison.}\subfig{c} shows that \Mname~outperforms bSpMM with on average $1.76\times$ speedup on neighbor aggregation and improves effective computation by 75.8\% on average. 
Our SGT technique can maximize the non-zero density of each non-zero tile and significantly reduce the number of non-zero tiles to be processed. 
However, bSpMM in cuSPARSE has to comply with the strict input sparse pattern (indicated in official API documentation~\cite{bSpMM_API}). For example, all rows in the arrays must have the same number of non-zero blocks. Thus, more redundant computations (on padding non-structural non-zero blocks) in bSpMM lead to inferior performance. 
We also notice that for SC datasets with a high average node degree and clustered node distribution, bSpMM would benefit more due to its usage of a larger block size of 32$\times$32 (fewer TCU invocations) compared to 16$\times$8 in TC-GNN ( more TCU invocations).

\textbf{Compared with tSparse and Triton}
From Table~\ref{tbl: Compare with tSparse, Triton}, \Mname~can outperform tSparse with on average $3.60\times$ speedup on SpMM. The major reason behind this is that \Mname~can well reduce the graph structural-level irregularity through our novel SGT strategy to benefit the dense TCU-based computation. In contrast, tSparse only considers partitioning the input sparse matrix into dense/sparse tiles based on their non-zero elements but ignores the potential of compressing non-zero elements into fewer tiles to reduce the workload.
\Mname~also outperforms Triton with on average $5.42\times$ speedup on SpMM. 
Triton's block-sparse GEMM for TCU acceleration is designed for dense neural networks (focusing on feature maps' sparsity), which is quite different from GNNs (focusing on the graph adjacency matrix's sparsity) with significantly larger sparse matrix size and more irregular pattern.
\begin{figure}[t] \small
    \centering
    \includegraphics[width=0.9\columnwidth]{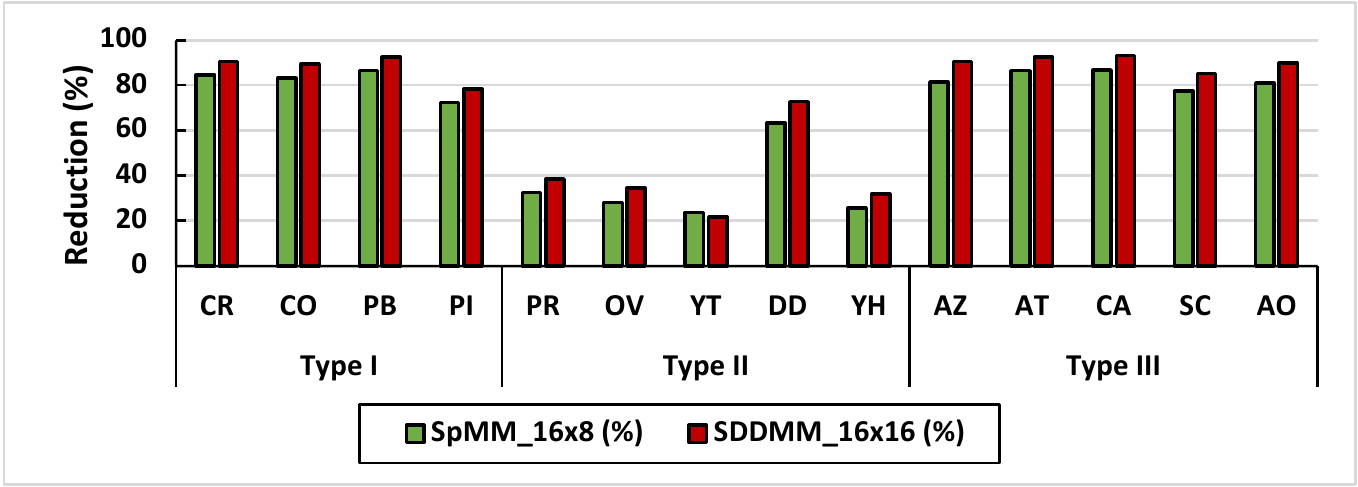}
    %
    \caption{SGT Effectiveness on SpMM and SDDMM.}
    \label{fig: SGT Effectiveness on SpMM and SDDMM}
\end{figure} 

\subsection{Additional Studies} 
\label{sect: Additional Studies}
\begin{figure}[t] \small
    \centering
    \includegraphics[width=0.8\columnwidth]{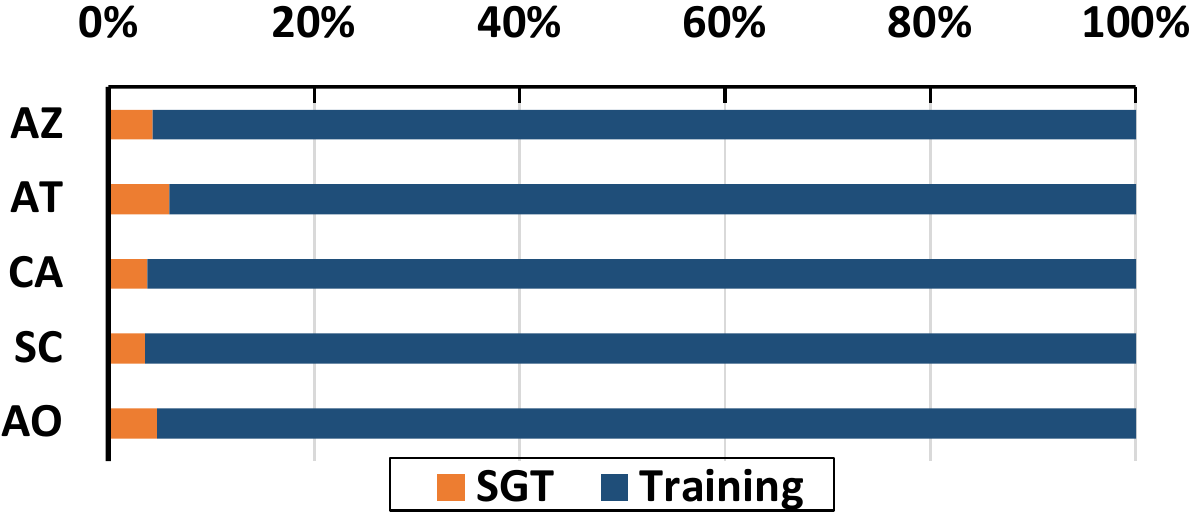}
    \caption{The overhead analysis of SGT.}
    \label{fig: The overhead analysis of SGT.}
\end{figure}

\textbf{SGT Effectiveness \& Overhead} 
We conduct a quantitive analysis of SGT in terms of the total number of TCU blocks between graphs w/o SGT and the graphs w/ SGT applied.
Note that in the SpMM-based aggregation, the size of TCU blocks is $16\times8$ since it serves as one of the operands in TCU GEMM. While in SDDMM-based edge feature computation, the size of TCU blocks is $16\times16$ since it serves as the resulting matrix of TCU GEMM.
Figure~\ref{fig: SGT Effectiveness on SpMM and SDDMM} shows that across all types of datasets, our SGT technique can significantly reduce the number of traversed TCU blocks (on average 67.47\%). 
The major reason is that SGT can largely improve the density of non-zero elements within each TCU block. In contrast, the graphs w/o SGT would demonstrate a large number of highly sparse TCU blocks. 
What is also worth noticing is that on Type II graphs, such a reduction benefit is lower. The reason is that Type II graphs consist of a set of small subgraphs that only maintain the intra-subgraph connections, which already maintain dense columns.
We evaluate the overhead of SGT (Figure~\ref{fig: The overhead analysis of SGT.}), we find that its overhead is consistently low (on average 4.43\%) compared with the overall training time (200 epoches as DGL~\cite{wang2019dgl}).

\begin{table}[t] \small
\caption{Sparsity Analysis. Numbers for bSpMM/TC-GNN are in GFLOPs. ``DB/W'': dense blocks per row window.}
\centering
\scalebox{0.95}{
\begin{tabular}{l|r r r}
\specialrule{.1em}{.05em}{.05em} 
\textbf{DB/W} & \textbf{Sparsity (\%)} & \textbf{bSpMM} & \textbf{TC-GNN} \\ \hline\hline
1 & 99.61 & 773.86 & 12,686.02 \\ 
2 & 99.22 & 1,597.83 & 11,010.75 \\ 
4 & 98.44 & 3,348.75 & 18,164.08 \\ 
8 & 96.88 & 6,528.10 & 25,883.10 \\ 
16 & 93.75 & 12,955.40 & 23,865.99 \\ 
32 & 87.50 & 26,061.70 & 16,629.28 \\ \specialrule{.1em}{.05em}{.05em} 
\end{tabular}}
\label{tbl: Sparsity Analysis}
\end{table}

\textbf{Sparsity Analysis} 
We compare with bSpMM on synthetic matrix data with different sparsity (zero-element ratio).
Note that we change the sparsity by varying the number of dense non-zero blocks (16$\times$16) within each row window, the input adjacent matrix size is fixed to 4096$\times$4096 while the dense embedding matrix dimension is fixed to 16. Table~\ref{tbl: Sparsity Analysis} shows that when sparsity increases from 93.75\% to 99.61\%, TC-GNN design demonstrates more throughput performance strength (averaged 6.9$\times$) and this is also the common sparsity range (more than 95\%) for most input graphs of GNNs. When the sparsity drops to around 87.50\% the sparse would demonstrate more advantage due to more dense blocks for computation.
\begin{figure}[t] \small
    \centering
    \includegraphics[width=0.7\columnwidth]{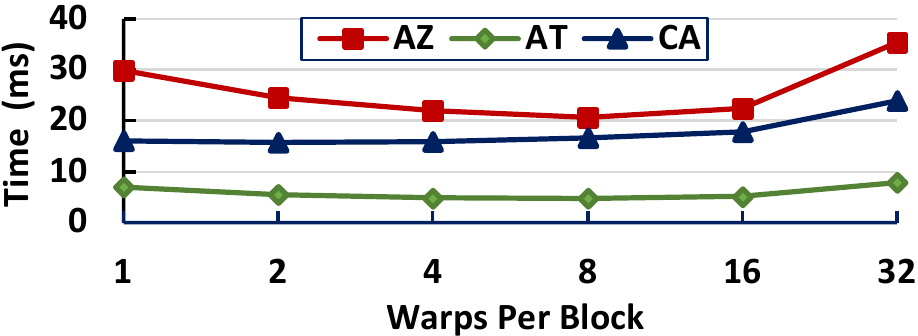}
    %
    \caption{Performance Impact of Warps per Block.}
    \label{fig: warp per block}
\end{figure} 

\textbf{Warps per Block: }
Figure~\ref{fig: warp per block} shows that with the increase of the number of warps, the overall performance for training per epoch would first decrease due to the better parallelism for loading the graph data. 
However, the number of warps per block would decrease the overall performance under certain circumstances (e.g., 32). 
All three settings suffer from evident performance degradation. 
Because the global memory access contention will become severe, leading to lower execution performance. 
Different datasets would have different ``optimal'' choices of the warp-per-block parameter. 
For example, on the CA dataset, 2 warps per block can deliver the best performance, while AZ requires 8 warps per block.  Based on our profiling and empirical study, the selection of this parameter should consider the average \textit{\#edges} per row window ($\mathit{avg.edges}$), which can be easily get during the preprocessing. Our \textbf{\texttt{preprocessor}} will generate $\mathit{warpPerBlock} = \lfloor \frac{\mathit{avg.edge}}{32}\rfloor$ to approach the ``optimal'' performance. For instance, the average edges per row window are 88 for CA, it reaches the best performance at 2 warps per block.

\begin{figure}[t] \small
    \centering
    \includegraphics[width=0.7\columnwidth]{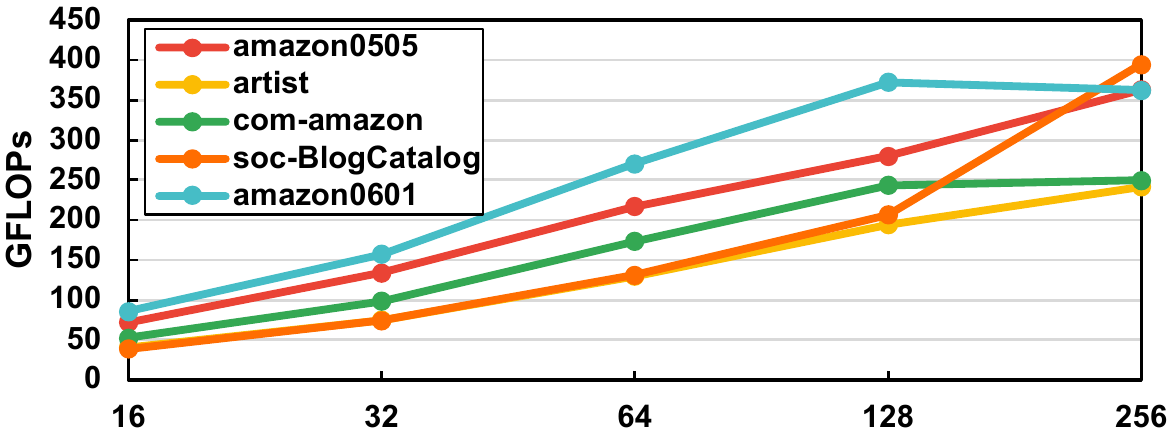}
    \caption{Analysis of \Mname~kernel throughput when increasing the node embedding dimension from 16 to 256.}
    \label{fig: Throughput analysis.}
\end{figure}
\textbf{Throughput Analysis:} For sparse matrix computations in GNNs, we measure the throughput performance of SpMM in \Mname~when the dimension of node embedding increases for a roofline analysis. Because sparse matrix computation is largely limited by its memory access performance, which is quite different from the dense GEMM computation that is largely bounded by the computing performance.
Figure~\ref{fig: Throughput analysis.} shows that the throughput of \Mname~can scale proportionally with the growing number of node embedding dimensions. 
This also indicates that \Mname~can effectively handle the  graphs with high-dimensional node embeddings and well utilize GPU resources. 
\section{Related Work and Discussion} 
\label{sect: Related Work}
{\textbf{Other GPUs} \Mname~can easily generalize to other GPUs (e.g., A6000, H100, and RTX4090) with TCUs via recompilation (\texttt{python setup.py install}).
TC-GNN also supports different TCU configurations (e.g., precision) by modifying (\textit{BLK\_H}, \textit{BLK\_W} in \texttt{TCGNN\_conv/config.h}) and four parameters (\texttt{M, N, K, dataType}) in \texttt{wmma::fragment}, then recompile.}
{For future GPUs with more TCUs, our \Mname~can also be adapted to accommodate such changes and maintain its performance advantage. There are two future GPU designs that we anticipate. The first direction is to place more TCUs per SM while keeping the total number of SMs unchanged. There will be more active warps per thread block (This is mainly because TCUs are operated by warps) and each warp will process fewer neighbors. The cost of decomposition and mapping can be offset by parallelism among more warps. The second direction is to place more SMs on GPUs while keeping TCUs per GPU  unchanged. In this scenario, there will be more thread blocks and each thread block will process neighbors from fewer nodes. The cost can be offset by parallelism among more thread blocks.}

\textbf{Other GNN Frameworks} Besides DGL and PyG, other single-GPU GNN frameworks like GNNAdvisor~\cite{GNNAdvisor}, GE-SpMM~\cite{ge-spmm}, and fuseGNN~\cite{chen2020fusegnn}, tailor their own GNN layers manually with low-level GPU kernel optimizations. 
Unfortunately, these designs limit their kernel optimizations to CUDA cores, thus, missing the golden opportunities to exploit the full potential of widely deployed AI-tailored GPUs with TCUs.

\textbf{Graph Partitioning/Reordering}
ROC~\cite{jia2020improving} introduces a learning-based graph partitioning to reduce the data movement between CPU and GPU when processing large graphs. 
%
Rabbit Order~\cite{arai2016rabbit} and Reverse Cuthill Mckee Algorithm~\cite{cuthill1969reducing} are focusing on \textit{row reordering/clustering} to improve node/row-wise computation locality.  
Our sparse-graph translation (SGT) technique is orthogonal and complementary to these graph partitioning and reordering techniques since our SGT focuses on \textit{column (neighbor) re-indexing} to improve neighbor-wise locality for TCU computation. 

\textbf{Distributed GNN Computation} There are two major ways of scaling-up GNN computing: 1) \textit{Distributed sampled graphs}~\cite{wang2019dgl, yang2022gnnlab, pyG, min2021pytorch} (where graph nodes and their embeddings are on the same GPU): \Mname~can be incorporated directly since all sampled graphs along with their node embeddings are presented at the same GPU. 
2) \textit{Distributed full-graph}~\cite{gandhi2021p3, ma2019neugraph, jia2020improving, wang2022empowering} (where graph nodes and their embeddings may be on different GPUs): \Mname~needs to be modified slightly by incorporating inter-GPU communication techniques (e.g., Unified Virtual Memory~\cite{unifiedMemory} and NVSHMEM~\cite{nvshmem}) to support the remote neighbor embedding access. We leave such exploration for our future work.

\section{Conclusion} 
\label{sect: conclusion}
In this paper, we introduce \Mname, the first GNN acceleration framework on TCU of GPUs.
We design a novel sparse graph translation technique to gracefully fit the sparse GNN workload on dense TCUs. 
Our TCU-tailored GPU kernel design maximizes the TCU performance gains for GNN computing through effective CUDA core and TCU collaboration and a set of memory/data flow optimizations.
Our seamless integration with the PyTorch framework further facilitates end-to-end GNN computing with high programmability.
Extensive experiments demonstrate the performance advantage of \Mname~over the state-of-the-art frameworks. 
across diverse GNN models and datasets.

Furthermore, our \Mname~design could also inspire potential TCU-like hardware features that can support (i) the dynamic shape of TCU input tiles and (ii) the dynamic structural sparsity of input tiles to yield higher performance benefits at the runtime. These proposed hardware features will help reduce more unnecessary computation in a more fine-grained and precise manner.

\section{Acknowledgment}
We would like to thank our shepherd, Asim Kadav, and the
anonymous USENIX ATC reviewers.
This work was supported in part by NSF-2124039 and CloudBank~\cite{norman2021cloudbank}. We also appreciate the generous help and support from NVIDIA Graduate Fellowship 2022-2023 for Yuke Wang and Amazon Faculty Award 2021-2022 for Yufei Ding.

\bibliographystyle{plain}
\bibliography{reference}


\clearpage
\appendix
\section{Artifact Appendix}

\textbf{\Mname} is the first TCU-based GNN acceleration design on GPUs.
\textit{\textbf{At the input level}}, \Mname~is equipped with a new \textit{sparse graph translation} (SGT) technique that can effectively identify those non-zero tiles and condense non-zero elements from these tiles into fewer number of ``dense'' tiles. 
\textit{\textbf{At the GPU kernel level}}, \Mname~exploits the benefits of CUDA core and TCU collaboration. The major design idea is that the CUDA core, which is more powerful at fine-grained thread-level execution, would be a good candidate for managing memory-intensive data access. While TCU, which is more powerful in handling simple arithmetic operations (\textit{e.g.}, multiplication and addition), would be well-suited for compute-intensive GEMM on dense tiles generated from SGT.
\textit{\textbf{At the framework level}},  \Mname~is integrated with the popular PyTorch framework to reduce extra learning efforts and improve user productivity and code portability.

\begin{itemize}
\itemsep0em 
    \item Code repository: \textbf{Github}\footnote{\url{https://github.com/YukeWang96/TC-GNN_ATC23.git}} and \textbf{Zenodo}\footnote{\url{https://doi.org/10.5281/zenodo.7893174}}.
    \item \textbf{Hardware, OS \& Compiler}:
    \begin{itemize}
    \itemsep0em 
    \item Intel Xeon Sliver 4110 CPU (8-core 16-threads) with 64GB host memory, NVIDIA RTX3090 GPU with 24 GB device memory.
    \item  Operating systems and versions: Ubuntu 16.04+.
    \item  Compilers and versions: NVCC-11.1+, GCC-7.5.0+
    Libraries and versions: CUDA-11.1+, Pytorch-1.8.0, DGL-v0.6.0, PyG-1.6.3 Input datasets and versions: SNAP network datasets.
\end{itemize}
\end{itemize}

\subsection*{Step-1: Environment Setup}
\subsubsection*{- 1.1a. [Method-1] Install via Docker (Recommended).}
\begin{lstlisting}[style=tt1]
cd docker/
./launch.sh
\end{lstlisting}

\noindent \textbf{- 1.1b. [Method-2] Install via Conda.}
\begin{lstlisting}[style=tt1]
curl -O https://repo.anaconda.com/archive/Anaconda3-2021.05-Linux-x86_64.sh
bash Anaconda3-2019.03-Linux-x86_64.sh
source ~/.bashrc
conda create -n env_name python=3.6
conda install pytorch torchvision torchaudio cudatoolkit=11.1 -c pytorch -c conda-forge
conda install -c dglteam dgl-cuda11.0
pip install torch requests tqdm
pip install torch-scatter -f https://pytorch-geometric.com/whl/torch-1.8.0+cu111.html
pip install torch-sparse -f https://pytorch-geometric.com/whl/torch-1.8.0+cu111.html
pip install torch-cluster -f https://pytorch-geometric.com/whl/torch-1.8.0+cu111.html
pip install torch-spline-conv -f https://pytorch-geometric.com/whl/torch-1.8.0+cu111.html
pip install torch-geometric
\end{lstlisting}

\noindent \textbf{- 1.2. Install TC-GNN.} 
\begin{lstlisting}[style=tt1]
cd TCGNN_conv/
./0_build_tcgnn.sh
\end{lstlisting}

\vspace{7pt}
\noindent \textbf{- 1.3. Download Datasets.}
\begin{lstlisting}[style=tt1]
wget https://storage.googleapis.com/graph_dataset/tcgnn-ae-graphs.tar.gz
tar -zxvf tcgnn-ae-graphs.tar.gz
rm -rf tcgnn-ae-graphs.tar.gz
\end{lstlisting}

\subsection*{Step-2. Run Major Experiments.}
\vspace{5pt}
\noindent \textbf{- 2.1. TC-GNN model End-to-End.}
\begin{lstlisting}[style=tt1]
./0_run_tcgnn_model.sh
\end{lstlisting}
Results: \texttt{1\_bench\_gcn.csv} \\
and \texttt{1\_bench\_agnn.csv}.

\vspace{7pt}
\noindent \textbf{- 2.2. DGL baseline (Fig-6a).}
\begin{lstlisting}[style=tt1]
cd dgl_baseline/ 
./0_run_dgl.sh
\end{lstlisting}
Results: \texttt{Fig\_6a\_dgl\_gcn.csv} \\
and \texttt{Fig\_6a\_dgl\_agnn.csv}.

\vspace{7pt}
\noindent \textbf{- 2.3. TC-GNN single kernel.}
\begin{lstlisting}[style=tt1]
./0_run_tcgnn_single_kernel.sh
\end{lstlisting}
Results: \texttt{1\_bench\_gcn.csv}
and \texttt{1\_bench\_agnn.csv}.

\vspace{7pt}
\noindent \textbf{- 2.4. cuSPARSE-bSpMM Baseline (Fig-6c).}
\begin{lstlisting}[style=tt1]
cd TCGNN-bSpmm/cusparse
./0_run_bSpMM.sh
\end{lstlisting}
Results: \texttt{Fig\_6c\_cuSPARSE\_bSpMM.csv}.

\vspace{7pt}
\noindent \textbf{- 2.5. Dense Tile Reduction (Fig-7)}.
\begin{lstlisting}[style=tt1]
python 3_cnt_TC_blk_SDDMM.py
python 3_cnt_TC_blk_SpMM.py
\end{lstlisting}
Results: \texttt{3\_cnt\_TC\_blk\_SDDMM.csv} \\ and  \texttt{3\_cnt\_TC\_blk\_SDDMM.csv}.

\vspace{7pt}
\noindent \textbf{- 2.6. tSparse Baseline (Table-5, column-2)}.
\begin{lstlisting}[style=tt1]
cd TCGNN-tsparse/
./0_run_tSparse.sh
\end{lstlisting}
Result: \texttt{Table\_5\_tSparse.csv}.

\vspace{7pt}
\noindent \textbf{- 2.7. Triton Baseline (Table-5, column-3).}
\begin{lstlisting}[style=tt1]
cd TCGNN-trition/python/bench
./0_run_triton.sh
\end{lstlisting}
Result: \texttt{1\_run\_triton.csv}.

\end{document}